\documentclass[11pt]{article}
\usepackage{CJKutf8}

\usepackage{tcolorbox}
\usepackage{listings}
\usepackage{enumitem}
\usepackage[]{acl}

\usepackage{amsmath,amsfonts,bm}
\usepackage{mathtools}
\usepackage{upgreek}

\def\eqref#1{equation~\ref{#1}}

\def\1{\bm{1}}

\DeclareMathAlphabet{\mathsfit}{\encodingdefault}{\sfdefault}{m}{sl}
\SetMathAlphabet{\mathsfit}{bold}{\encodingdefault}{\sfdefault}{bx}{n}

\usepackage{amsthm}

\usepackage{times}
\usepackage{latexsym}
\usepackage{color,xcolor}
\usepackage{tikz}
\usepackage{multirow}
\usepackage{xspace}
\usepackage{booktabs}
\usepackage{ulem}
\usepackage{graphicx} 
\usepackage{caption}
\usepackage{tcolorbox}
\usepackage{fvextra}
\usepackage{physics}
\usepackage{wrapfig}
\usepackage{bbm}
\usepackage[flushleft]{threeparttable}
\usepackage{verbatim}
\usepackage{scalerel} 
\usepackage{microtype}
\usepackage{amssymb}
\usepackage{utfsym}
\usepackage{pifont}
\usepackage{svg}
\usepackage{hyperref}
\hypersetup{
    colorlinks=true,
    linkcolor=red,
    citecolor=blue,
    filecolor=magenta,      
    urlcolor=blue,
linktocpage}
\usepackage{url}
\usepackage{enumitem}
\usepackage{tabularx}
\usepackage{courier}
\usepackage{makecell}
\usepackage{amsmath}
\usepackage{algorithm}
\usepackage{algpseudocode}
\usepackage{listings}
\usepackage{xcolor}
\usepackage{fancyvrb}
\usepackage{color}
\usepackage{colortbl}
\definecolor{color_1}{HTML}{B8D9EB}
\definecolor{color_2}{HTML}{FEFF92}
\definecolor{color_3}{HTML}{99F0C2}
\usepackage{highlight}

\usepackage{subcaption}
\renewcommand{\eqref}[1]{Eq.~\ref{#1}}

\NewDocumentCommand{\hao}
{ mO{} }{\textcolor{blue}{\textsuperscript{\textit{hao}}\textsf{\textbf{\small[#1]}}}}

\NewDocumentCommand{\heng}
{ mO{} }{\textcolor{red}{\textsuperscript{\textit{Heng}}\textsf{\textbf{\small[#1]}}}}

\NewDocumentCommand{\jiateng}
{ mO{} }{\textcolor{purple}{\textsuperscript{\textit{jiateng}}\textsf{\textbf{\small[#1]}}}}
\NewDocumentCommand{\yy}
{ mO{} }{\textcolor{yellow}{\textsuperscript{\textit{yy}}\textsf{\textbf{\small[#1]}}}}
\NewDocumentCommand{\xingyao}
{ mO{} }{\textcolor{orange}{\textsuperscript{\textit{xingyao}}\textsf{\textbf{\small[#1]}}}}
\NewDocumentCommand{\zihan}
{ mO{} }{\textcolor{green}{\textsuperscript{\textit{zihan}}\textsf{\textbf{\small[#1]}}}}

\newcommand{\ourbench}[0]{ConceptMath\xspace}

\usepackage[utf8]{inputenc}

\usepackage{longtable}
\usepackage{multirow}
\usepackage{booktabs}

\usepackage[utf8]{inputenc}

\title{ConceptMath: A Bilingual Concept-wise Benchmark for Measuring Mathematical Reasoning of Large Language Models}
\author{Yanan Wu*$^{1}$, Jie Liu*$^{2}$, Xingyuan Bu*$^{1}$, Jiaheng Liu$^{\dagger, 1}$, 
\textbf{Zhanhui Zhou}$^{1}$, \\  \textbf{Yuanxing Zhang}$^{1}$, \textbf{Chenchen Zhang}$^{1}$,  \textbf{Zhiqi Bai}$^{1}$, \textbf{Haibin Chen}$^{1}$,  \textbf{Tiezheng Ge}$^{1}$, 
\\\textbf{Wanli Ouyang}$^{3}$, \textbf{Wenbo Su}$^{1}$, \textbf{Bo Zheng}$^{1}$\\
       $^{1}$Alibaba Group; $^{2}$The Chinese University of Hong Kong; \\
       $^3$ Shanghai Artificial Intelligence Laboratory \\
        \texttt{\{lixing.wyn, ljh411989\}@taobao.com}}

\begin{document}

        \maketitle
                \let\thefootnote\relax\footnotetext{* First three authors contributed equally.}
                \let\thefootnote\relax\footnotetext{$^\dagger$ Corresponding Author: Jiaheng Liu.}
\begin{abstract}

This paper introduces \ourbench, a bilingual (English and Chinese), fine-grained benchmark that evaluates concept-wise mathematical reasoning of Large Language Models (LLMs). 
Unlike traditional benchmarks that evaluate general mathematical reasoning with an average accuracy, 
\ourbench systematically organizes math problems under a hierarchy of math concepts,
so that mathematical reasoning can be evaluated at different granularity with concept-wise accuracies.
Based on our ConcepthMath,
we evaluate a broad range of LLMs,
and we observe existing LLMs, though achieving high average accuracies on traditional benchmarks, exhibit significant performance variations across different math concepts and may even fail catastrophically on the most basic ones.
Besides,
we also introduce an efficient fine-tuning strategy to enhance the weaknesses of existing LLMs.
Finally, we hope \ourbench could guide the developers to understand the fine-grained mathematical abilities of their models and facilitate the growth of foundation models\footnote{The data and code are available at \url{https://github.com/conceptmath/conceptmath}.}.
\end{abstract}

\section{Introduction}

\begin{figure}[t]
    \centering
    \includegraphics[width=1.0\linewidth]{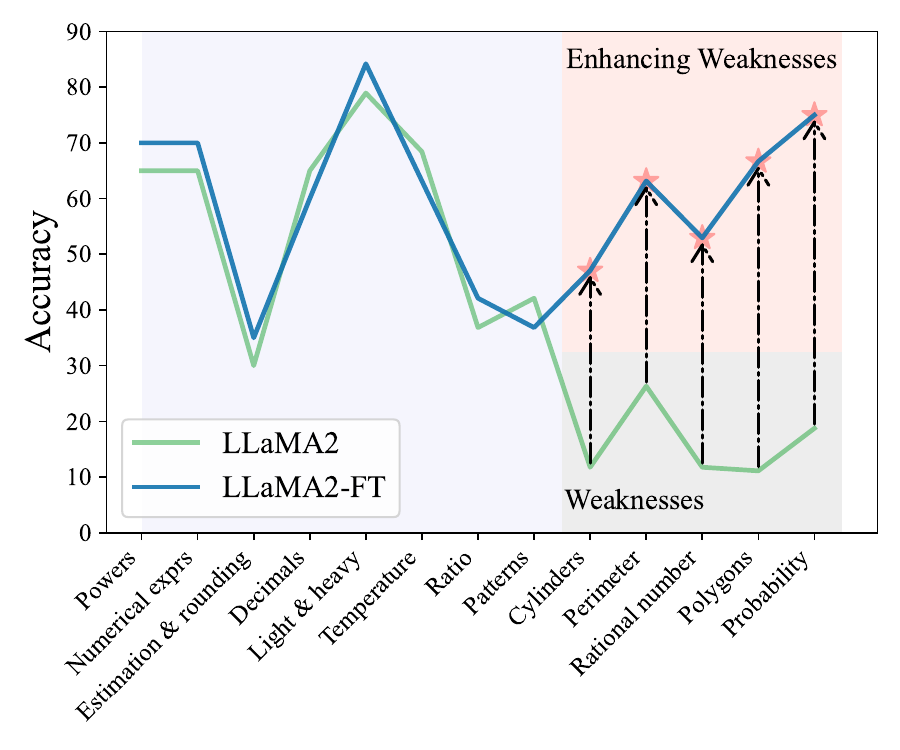}
    \vspace{-6mm}
    \caption{The concept-wise
    accuracies of LLaMA2-13B and the fine-tuned version based on our efficient fine-tuning method (i.e., LLaMA2-FT).}
    \label{fig: motivation}
    \vspace{-3mm}
\end{figure}

Mathematical reasoning is a crucial capability for Large Language Models (LLMs).
Recent advancements in LLMs, including Anthropic~\cite{anthropic2023claude}, GPT-4~\citep{gpt4}, and LLaMA~\citep{touvron2023llama}, have demonstrated impressive mathematical reasoning on existing benchmarks with high average accuracies on datasets like GSM8K~\citep{cobbe2021training}.
Although these benchmarks are able to measure the overall mathematical reasoning capabilities of LLMs \textit{on average}, they fail to probe the fine-grained failure modes of mathematical reasoning \textit{on specific mathematical concepts}.
For example, 
Fig.~\ref{fig: motivation} shows that the performance of LLaMA2-13B varies significantly across different concepts and fails on simple concepts like \textit{Rational number} and \textit{Cylinders}.
It is crucial to know these specific failure modes of the language model, especially
in some practical applications where
we need to focus on specific mathematical abilities.
For example,
for financial analysts,
calculation and statistics are the concepts of most interest while others like geometry are not as important. 

Moreover, the mathematics system, by its nature, is more fine-grained than holistic. It is typically organized into distinct math concepts~\footnote{\url{https://en.wikipedia.org/wiki/Lists_of_mathematics_topics}}, and humans develop comprehensive mathematical capabilities through a concept-by-concept, curriculum-based learning process~\citep{simon2011studying, fritz2013development}. These issues underscore the core motivation of
this paper: \textit{the need for a fine-grained benchmark that evaluates concept-wise mathematical reasoning capabilities of LLMs}.

\begin{figure*}[t!]
  \centering
\begin{subfigure}[b]{0.49\textwidth}
\includegraphics[width=0.91\linewidth]{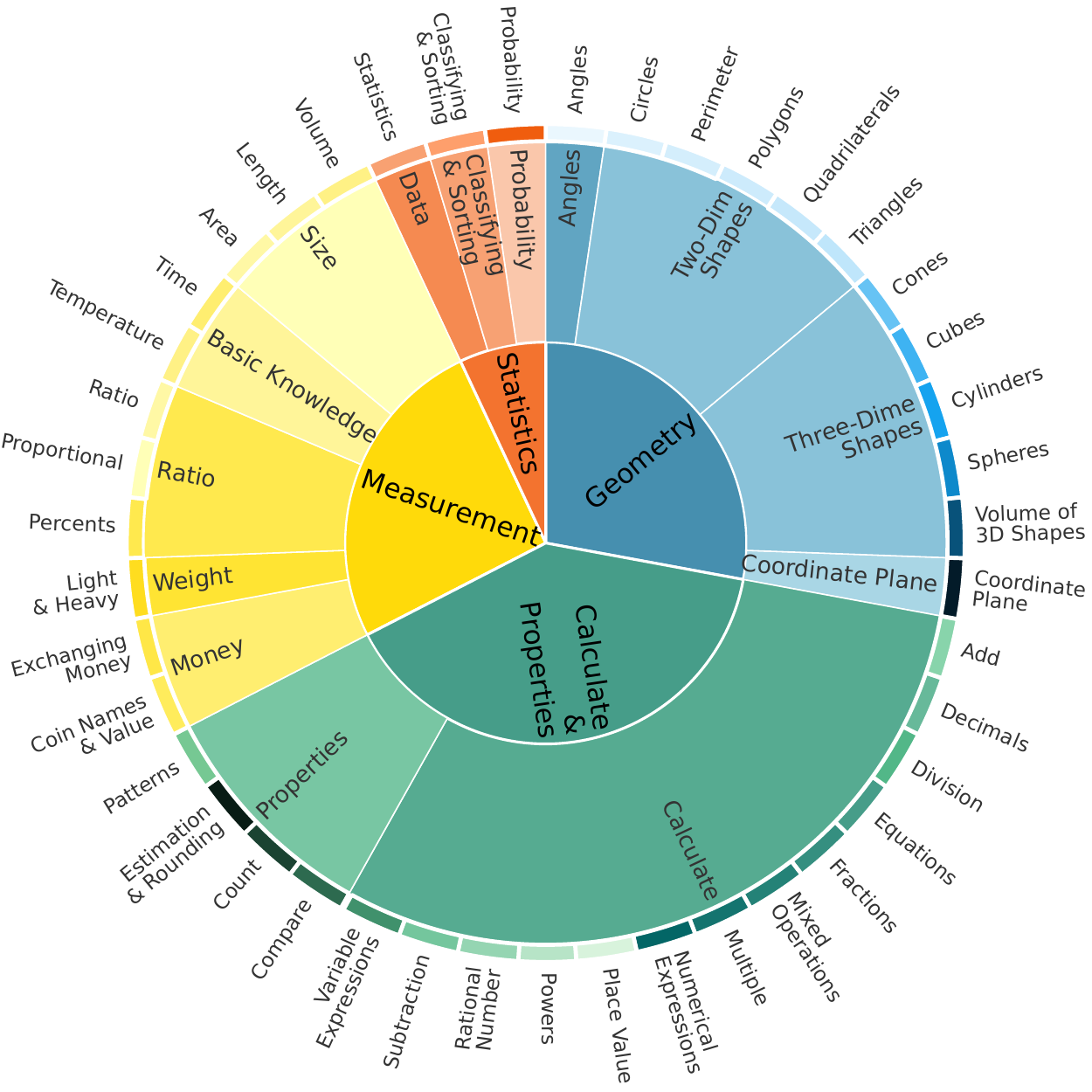}
    \caption{English Elementary (Elementary-EN)}
\end{subfigure}
\begin{subfigure}[b]{0.49\textwidth}
    \includegraphics[width=1.0\linewidth]{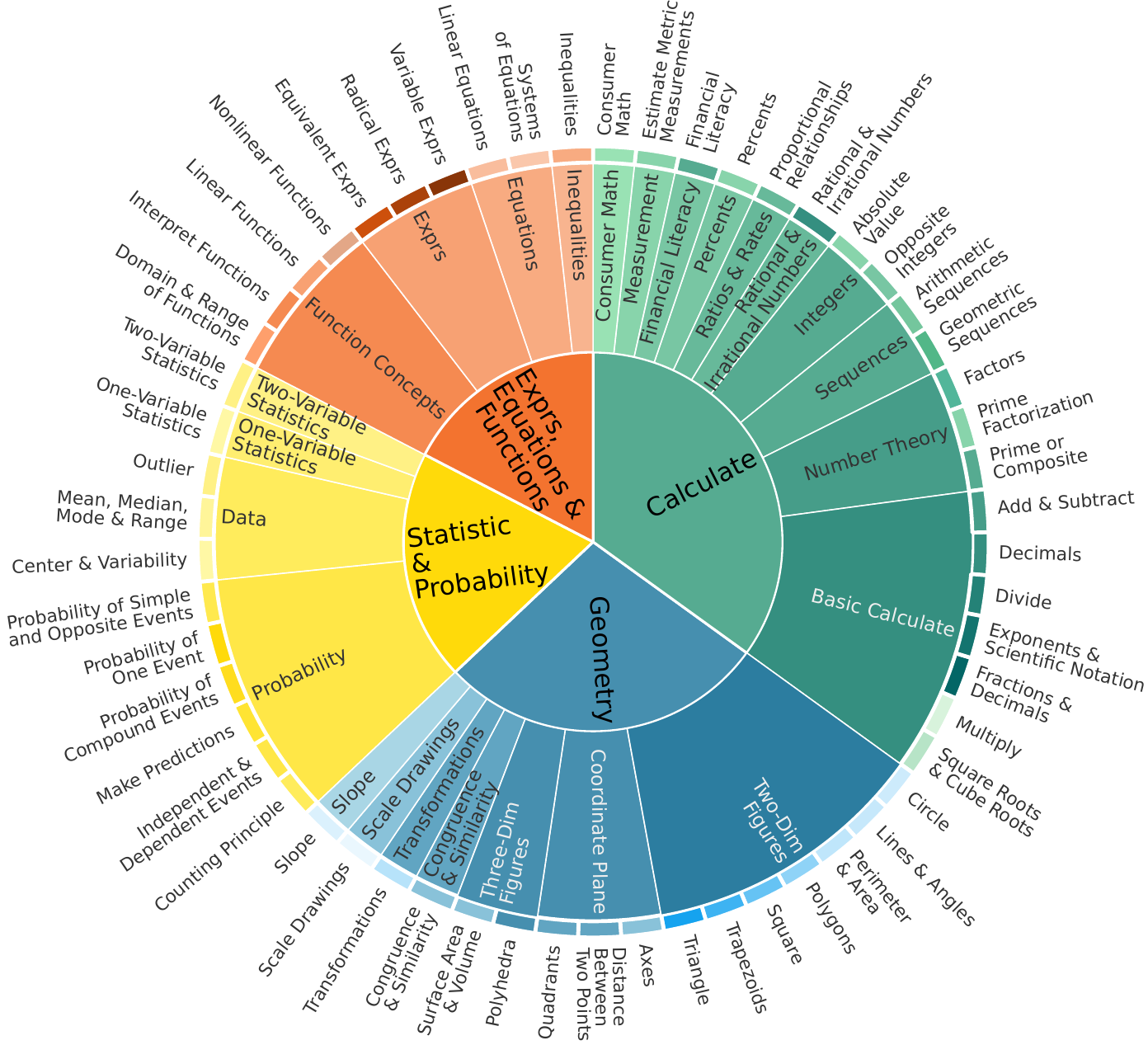}
    \caption{English Middle (Middle-EN)}
\end{subfigure}
\begin{subfigure}[b]{0.49\textwidth}
    \includegraphics[width=0.92\linewidth]{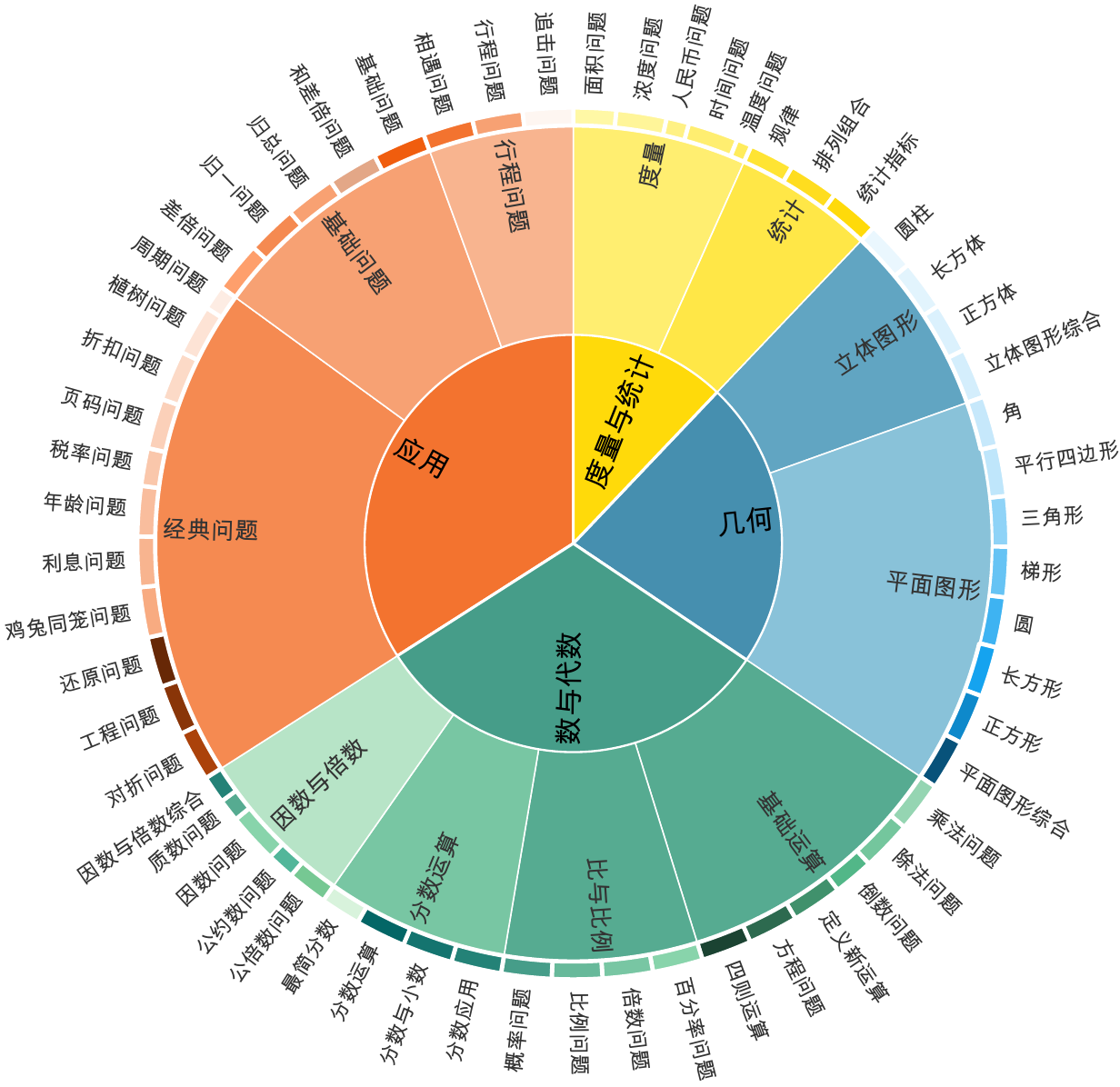}
    \caption{Chinese Elementary (Elementary-ZH)}
\end{subfigure}
\begin{subfigure}[b]{0.49\textwidth}
    \includegraphics[width=1.0\linewidth]
    {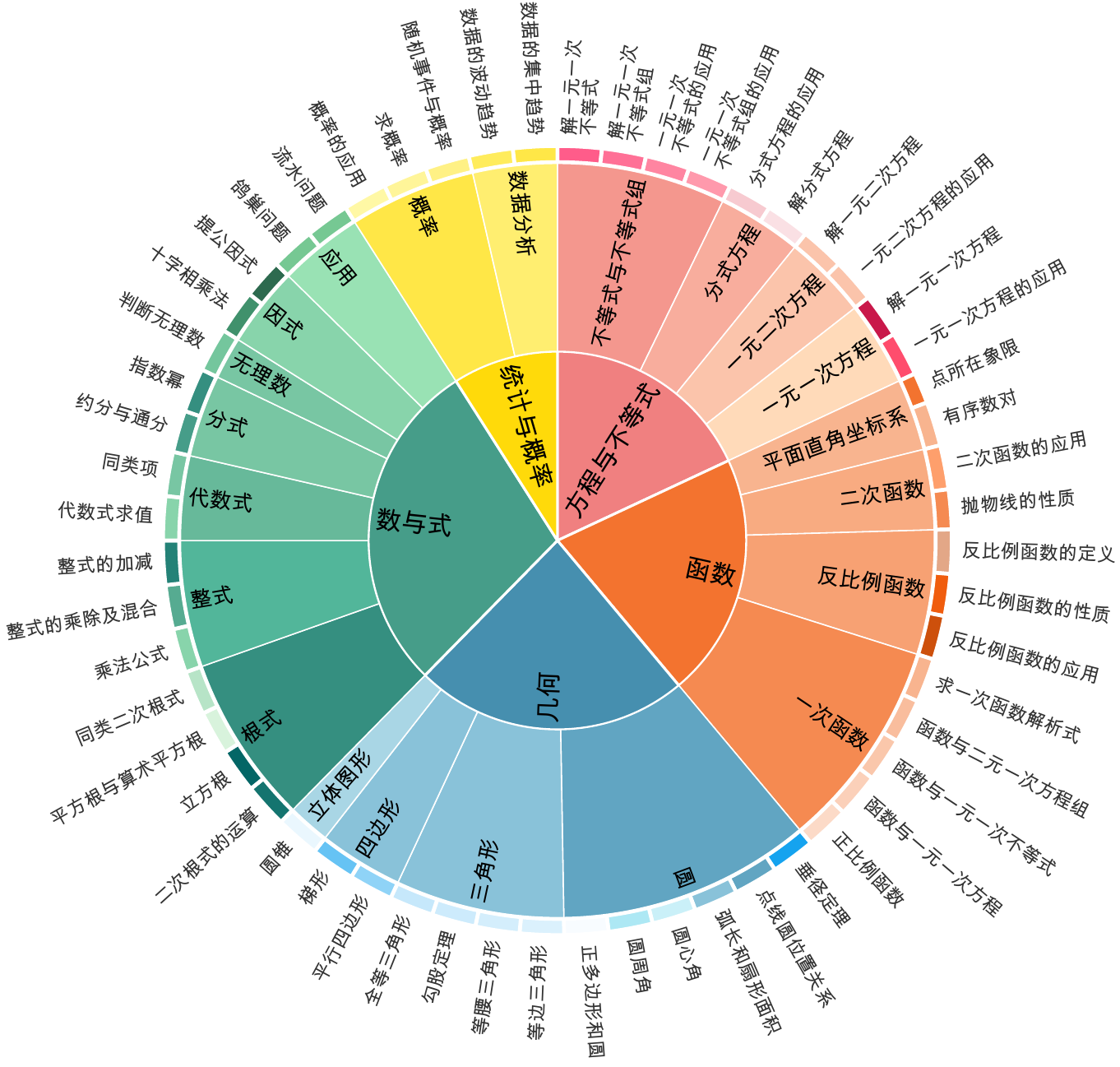}
    \caption{Chinese Middle (Middle-ZH)}
\end{subfigure}
  \caption{Diagram overview of four concept systems in \ourbench. We have provided translated Chinese concept names in English (See Appendix~\ref{app: concept}).}
 \label{fig:diagram-overview}
 \vspace{-2mm}
\end{figure*}

Therefore, first,
we introduce \ourbench, the first bilingual (English and Chinese), concept-wise benchmark for measuring mathematical reasoning. 
\ourbench gathers math concepts from four educational systems, resulting in four distinct mathematical concept systems: \textit{English Elementary}, \textit{English Middle}, \textit{Chinese Elementary}, and \textit{Chinese Middle}~\footnote{The four concept systems are abbreviated as \textbf{Elementary-EN}, \textbf{Middle-EN}, \textbf{Elementary-ZH}, and \textbf{Middle-ZH}.}.
Each of these concept systems organizes around 50 atomic math concepts under a three-level hierarchy and each concept includes approximately 20 mathematical problems.
Overall, \ourbench comprises a total of 4011 math word problems across 214 math concepts,
and Fig.~\ref{fig:diagram-overview} shows the diagram overview of \ourbench.

Second, based on our ConceptMath,
we perform extensive experiments to assess the mathematical reasoning of existing LLMs, including 2 close-sourced LLMs and 17 open-sourced LLMs.
These evaluations were performed in zero-shot, chain-of-thought (CoT), and few-shot settings.
To our surprise, even though most of the evaluated LLMs claim to achieve high average accuracies on traditional mathematical benchmarks (e.g., GSM8K), they fail catastrophically across a wide spectrum of mathematical concepts. 

Third, to make targeted improvements on underperformed math concepts,
we propose an efficient fine-tuning strategy by first training a concept classifier and then crawling a set of samples from a large open-sourced math dataset~\cite{paster2023openwebmath,wang2023mathpile} for further LLMs fine-tuning.
In Fig.~\ref{fig: motivation},
for LLaMA2-FT,
we observe that the results of these weaknesses improved a lot after using the efficient fine-tuning method.

In summary, our contributions are as follows:
\begin{itemize}[leftmargin=4mm]
\vspace{-1mm}
    \item We introduce \ourbench, the first bilingual, concept-wise benchmark for measuring mathematical reasoning. \ourbench encompasses 4 systems, approximately 214 math concepts, and 4011 math word problems, which can guide further improvements on the mathematical reasoning of existing models.
    \vspace{-1mm}
    \item Based on \ourbench, we evaluate many LLMs and perform a comprehensive analysis of their results. For example, we observe that most of these LLMs (including open-sourced, closed-sourced, general-purpose, or math-specialized models) show significant variations in their performance results across math concepts.
\vspace{-1mm}
    \item We also evaluate the contamination rate of our ConceptMath and introduce a simple and efficient fine-tuning method to improve the weaknesses of existing LLMs.
\end{itemize}

\section{ConceptMath}

\ourbench is the first bilingual, concept-wise benchmark for measuring mathematical reasoning.
In this section, we describe the design principle, dataset collection process, dataset statistics and an efficient fine-tuning strategy to enhance the weaknesses identified by our ConceptMath.

\subsection{Design Principle}
We created \ourbench based on the following two high-level design principles:

\paragraph{Concept-wised Hierarchical System.} 
The primary goal of \ourbench is to evaluate the mathematical reasoning capacities of language models at different granularity. Therefore, \ourbench organizes math problems within a three-level hierarchy of mathematical concepts in Fig.~\ref{fig:diagram-overview}.
This approach provides concept-wise evaluation for mathematical reasoning of language models and makes targeted and effective improvements possible.

\paragraph{Bilingualism.}
Most of the current mathematical benchmark focuses solely on English, leaving multi-lingual mathematical reasoning unexplored. As an early effort to explore multi-lingual mathematical reasoning, we evaluate mathematical reasoning in two languages: English and Chinese. Besides, since cultures and educational systems vary across different languages, common math concepts can differ a lot. Therefore, we carefully collect concepts in both languages, instead of merely translating from one language to another.
For example,
measurement metrics (e.g., money, size) are different for English and Chinese.

\subsection{Data Collection}
Subsequently, for data collection, we take a two-step approach to operationalize the aforementioned design principles: First, we recruit experts to delineate a hierarchy of math concepts based on different education systems. Secondly, we collect problems for each concept from various sources or design problems manually, which is succeeded by quality assessment and data cleaning.

\paragraph{Math Concept System Construction.}
Since the education systems provide a natural hierarchy of math concepts, we recruited four teachers from elementary and middle schools, specializing in either English or Chinese, to organize a hierarchy of math concepts for different education systems. 
This leads to four concept systems: Elementary-EN, Middle-EN, Elementary-ZH, and Middle-ZH, with 
each system consisting of a three-level hierarchy of around 50 atomic math concepts (Fig.~\ref{fig:diagram-overview}).

\paragraph{Math Problem Construction.}

Then we conducted a thorough data acquisition from various sources (including educational websites, textbooks, and search engines with specific concepts) to collect math word problems (including both questions and answers) for each math concept.
To guarantee a balance across all concepts, approximately 20 problems were gathered for each math concept. Following this, both GPT-4~\cite{gpt4} and human experts were employed to verify and rectify the categorization and the solution of each problem.
However, we observed that for some concepts, the problem count was significantly below 20. To address this issue, manual efforts were undertaken to augment these categories, ensuring a consistent collection of 20 problems for each concept. Furthermore, to broaden the diversity of the dataset and minimize the risk of data contamination, all gathered problems were paraphrased using GPT-4.
It is important to note that the collection and annotation processes were carried out by a team of six members, each possessing a university degree in an engineering discipline, to maintain a high level of technical expertise in executing these tasks.

\subsection{Dataset Statistics}
\noindent\textbf{Comparison to existing datasets}. As shown in Table~\ref{tab:dataset_compare}, our ConceptMath differs from related datasets in
various aspects: (1) ConceptMath is the first dataset to study fine-grained mathematical concepts and encompasses 4 systems, 214 math concepts, and 4011 math word problems. 
(2) Problems in ConcepthMath are carefully annotated based on the mainstream education systems for English (EN) and Chinese (ZH). 

\noindent\textbf{Details on the hierarchical system}.
Apart from Fig.~\ref{fig:diagram-overview},
we also provide the details on the hierarchical system more clearly in Appendix~\ref{app: concept}. 

\begin{figure}[t]
    \centering
    \includegraphics[width=1.0\linewidth]{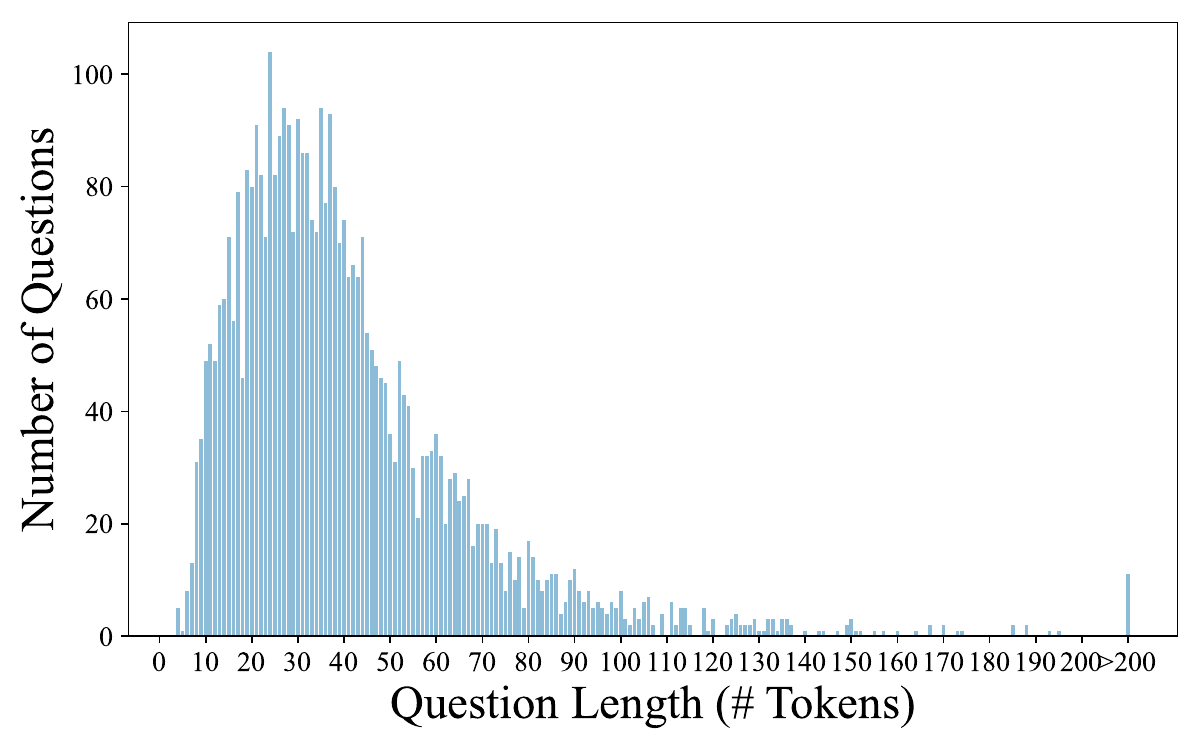}
    \vspace{-5mm}
    \caption{Length distributions of our ConceptMath.}
    \label{fig: length}
    \vspace{-3mm}
\end{figure}

\noindent\textbf{Length distribution.}
Fig.~\ref{fig: length} shows the length distribution of our ConcepthMath,
where number of tokens is reported~\footnote{We use the ``cl100k\_base'' tokenizer from \url{https://github.com/openai/tiktoken}}. 
 The minimum, average and maximum of the tokens for these questions are 4, 41 and 309, respectively,
 which shows that they have lexical richness.
\begin{table}[t!]
\centering
\setlength\tabcolsep{10pt}
    \resizebox{1\linewidth}{!}{
\begin{tabular}{c|ccc}
\toprule
Benchmark                                                 & Language                    & Fine-grained &Size                   \\
\midrule
GSM8K                     & EN                                       & \ding{55}                          & 1319                   \\
MATH                    &EN                                             & \ding{55}                         & 5000                  \\
TabMWP                                                             &EN& \ding{55}                          & 7686                  \\
Dolphin18K                &EN                       & \ding{55}                                                           & 1504                \\
Math23K                   &     ZH                  & \ding{55}                                                           & 1000               \\
ASDiv                     & EN                      & \ding{55}                                                            & 2305                   \\
SVAMP                     & EN                      & \ding{55}                                                            & 300                  \\ 
SingleOp &EN& \ding{55}&159\\
MMLU-Math&EN& \ding{55}&906\\
\midrule
ConceptMath                               & EN\&ZH                          & \ding{51}                           & 4011                   \\ 
\bottomrule
\end{tabular}
}
\vspace{-2mm}
\caption{A comparison of our ConceptMath with some notable mathematical datasets. Note that the size is the number of samples of the test split.}
\label{tab:dataset_compare}
\vspace{-4mm}
\end{table}

\subsection{Efficient Fine-Tuning}
Based on our \ourbench,
we are able to identify the weaknesses in the mathematical reasoning capability of LLMs through concept-wise evaluation. In this section, we explore a straightforward approach to enhance mathematical abilities towards specific concepts by first training a concept classifier and then curating a set of samples from a large open-sourced math dataset.
Specifically,
first, by additionally collecting {extra 10 problems per concept}, we construct a classifier capable of identifying the concept class of a given question. The backbone of this classifier is a pretrained bilingual LLM, where the classification head is operated on its last hidden output feature.
Then,
we proceed to fine-tune LLMs using this specific dataset combined with the existing general math dataset,
which aims to avoid overfitting on a relatively small dataset.
More details have been provided in the Appendix~\ref{app: eft}.

\section{Experiments}
In this section, we perform extensive experiments to demonstrate the effect of our \ourbench.
\vspace{-1mm}
\subsection{Experimental Setup}
\paragraph{Evaluated Models.}
We assess the mathematical reasoning of existing advanced LLMs on \ourbench, including 2 close-sourced LLMs (i.e., GPT-3.5/GPT-4~\citep{gpt4}) and 17 open-sourced LLMs (i.e., WizardMath-13B~\cite{luo2023wizardmath}, MetaMath-13B~\cite{yu2023metamath}, MAmmoTH-13B~\cite{yue2023mammoth}, Qwen-14B/72B~\cite{qwen}, Baichuan2-13B~\cite{baichuan}, ChatGLM3-6B~\cite{du2022glm}, InternLM2-7B/20B~\cite{2023internlm}, InternLM2-Math-7B/20B~\cite{ying2024internlmmath}, LLaMA2-7B/13B/70B~\cite{touvron2023llama2}, Yi-6B/34B~\cite{2023yi} and DeepSeekMath-7B~\cite{deepseek-math}). 
Note that WizardMath-13B, MetaMath-13B, and MAmmoTH-13B are specialized math language models fine-tuned from LLaMA2. InternLM2-Math and  DeepSeekMath-7B are specialized math language models fine-tuned from corresponding language models.
More details of these evaluated models can be seen in Appendix~\ref{app: em}.

\begin{table*}[]
\resizebox{1.0\textwidth}{!}{
\begin{tabular}{c|cccccccccccc|c}
\toprule
\multirow{2}{*}{\textbf{Model}} & \multicolumn{3}{c}{\textbf{Elementary-EN}} & \multicolumn{3}{c}{\textbf{Middle-EN}} & \multicolumn{3}{c}{\textbf{Elementary-ZH}} & \multicolumn{3}{c|}{\textbf{Middle-ZH}} & \multirow{2}{*}{\textbf{Avg.}} \\ \cmidrule(lr){2-4} \cmidrule(lr){5-7} \cmidrule(lr){8-10} \cmidrule(lr){11-13} 
                       & ZS    & ZS-COT       & FS    & ZS    & ZS-COT     & FS   & ZS     & ZS-COT        & FS    & ZS   & ZS-COT      & FS  &                          \\ \midrule
Yi-6B & 67.94  & 67.56  & 59.03  & 65.55  & 64.59  & 56.05  & 34.33  & 31.91  & 37.86  & 36.46  & 36.19  & 36.46  & 49.49  \\
ChatGLM3-6B & 60.69  & 63.10  & 53.18  & 51.25  & 60.17  & 51.34  & 46.23  & 43.63  & 40.74  & 44.77  & 43.32  & 40.43  & 49.90  \\
DeepSeekMath-7B & 66.92  & 77.35  & 73.92  & 56.53  & 69.87  & 66.31  & 60.47  & 62.33  & 64.19  & 56.50  & 56.95  & 56.86  & 64.02  \\
InternLM2-Math-7B & 71.12  & 72.01  & 69.59  & 63.44  & 62.96  & 63.05  & 57.30  & 58.23  & 58.60  & 53.79  & 53.16  & 53.88  & 61.43  \\
InternLM2-7B & 68.83  & 69.97  & 66.67  & 37.04  & 65.83  & 55.47  & 47.63  & 49.02  & 53.02  & 45.22  & 45.40  & 44.86  & 54.08  \\
LLaMA2-7B & 36.51  & 42.62  & 38.68  & 34.26  & 39.16  & 33.69  & 15.72  & 17.67  & 17.58  & 30.87  & 32.22  & 27.80  & 30.57  \\
\midrule
MAmmoTH-13B & 61.32  & 52.42  & 56.49  & 53.93  & 45.20  & 48.08  & 22.33  & 33.30  & 23.81  & 27.98  & 43.05  & 29.15  & 41.42  \\
WizardMath-13B & 41.73  & 44.78  & 34.99  & 36.85  & 37.72  & 45.11  & 10.51  & 11.26  & 18.70  & 12.36  & 15.52  & 22.92  & 27.70  \\
MetaMath-13B & 54.45  & 51.78  & 47.96  & 44.24  & 43.47  & 47.50  & 11.44  & 17.30  & 27.53  & 21.21  & 26.08  & 29.60  & 35.21  \\
Baichuan2-13B & 68.83  & 68.58  & 54.07  & 67.66  & 69.67  & 40.40  & 57.02  & 58.23  & 22.05  & 55.05  & 55.32  & 26.90  & 53.65  \\
LLaMA2-13B & 44.02  & 49.75  & 47.07  & 44.72  & 46.45  & 43.09  & 20.19  & 24.19  & 22.14  & 33.30  & 35.38  & 26.17  & 36.37  \\
Qwen-14B & 46.95  & 65.78  & 72.65  & 38.48  & 59.60  & 67.85  & 28.09  & 65.12  & 64.47  & 22.92  & 58.30  & 62.09  & 54.36  \\
\midrule
InternLM2-Math-20B & 74.05  & 75.32  & 73.41  & 64.11  & 71.21  & 70.83  & 62.98  & 61.95  & 61.77  & 55.14  & 55.78  & 56.86  & 65.28  \\
InternLM2-20B & 53.31  & 72.52  & 73.28  & 45.11  & 67.47  & 56.72  & 48.19  & 55.53  & 59.81  & 45.13  & 50.63  & 56.68  & 57.03  \\
Yi-34B & 74.68  & 73.66  & 56.36  & 72.26  & 74.66  & 65.83  & 50.05  & 51.16  & 38.79  & 45.40  & 43.95  & 40.97  & 57.31  \\
LLaMA2-70B & 56.11  & 60.31  & 30.53  & 58.06  & 60.94  & 31.67  & 28.65  & 26.70  & 24.37  & 37.64  & 34.30  & 28.43  & 39.81  \\
Qwen-72B & 77.10  & 75.06  & 77.23  & 74.66  & 69.87  & 73.99  & 71.16  & 68.65  & 61.86  & 71.30  & 65.43  & 62.45  & 70.73  \\
\midrule
GPT-3.5 & 85.75  & 92.37  & 84.35  & 83.88  & 90.12  & 82.73  & 56.47  & 53.21  & 56.93  & 51.90  & 53.52  & 55.69  & 70.58  \\
GPT-4 & 86.77  & 90.20  & 89.57  & 84.26  & 89.83  & 88.68  & 67.91  & 72.28  & 72.00  & 63.81  & 64.26  & 66.61  & 78.02  \\
\midrule
\textbf{Avg.} & 63.00  & 66.59  & 61.00  & 56.65  & 62.57  & 57.28  & 41.93  & 45.35  & 43.49  & 42.67  & 45.72  & 43.41  & 52.47 \\
\bottomrule
\end{tabular}}
\vspace{-2mm}
\caption{Results of different models on our constructed \ourbench benchmark dataset. Note that ``ZS'', ``ZS-COT'', ``FS'' represents ``zero-shot'', ``zero-shot w/ chain-of-thought'' and ``few-shot'', repsectively. Models are grouped roughly according to their model sizes.}
\vspace{-3mm}
\label{tab:mainresults}
\end{table*}

\paragraph{Evaluation Settings.} We employ three distinct evaluation settings: zero-shot, zero-shot with chain-of-thought (CoT), and few-shot promptings. The zero-shot prompting assesses the models' intrinsic problem-solving abilities without any prior examples. The zero-shot with CoT prompting evaluates the models' ability to employ a logical chain of thought. In the few-shot prompting setting, the model is provided with fixed 5-shot prompts for different systems  (See Appendix~\ref{app: prompts}), which includes five newly created examples with concise ground truth targets. This approach is designed to measure the in-context learning abilities. 
Besides, following MATH~\citep{hendrycks2021measuring}, all questions and answers in \ourbench have been carefully curated,
and each problem is evaluated based on exact matches.
Moreover, greedy decoding with a temperature of 0 is used.

\subsection{Results}

\begin{figure}[t!]
    \centering
    \includegraphics[width=1.0\linewidth]{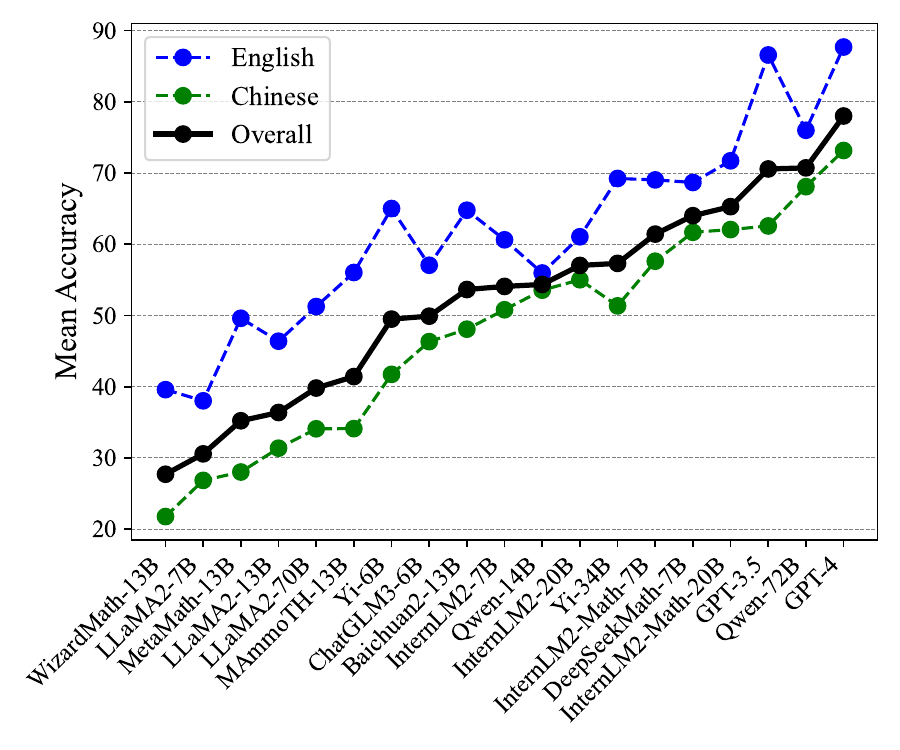}
    \vspace{-1mm}
    \caption{Mean accuracies for English, Chinese, and overall educational systems. }
    \vspace{-3mm}
    \label{fig: mean_acc}
\end{figure}

\paragraph{Overall Accuracy}
We present the overall accuracies of different LLMs on our \ourbench benchmark under various prompt settings in Table~\ref{tab:mainresults}. Subsequently, we analyzed the mathematical abilities of these LLMs in both English and Chinese in Fig.~\ref{fig: mean_acc}. Our analysis led to the following key findings:
(1) GPT-3.5/4 showcases the most advanced mathematical reasoning abilities among LLMs in both English and Chinese systems, and the leading open-source Qwen-72B model archives comparable performance compared with GPT-3.5. (2) The scores on Chinese systems of most existing LLMs are lower than English systems a lot. For example, accuracies on Middle-ZH and Middle-EN for GPT-4 are 63.81 and 84.26. (3) Several models (e.g., WizardMath-13B or MetaMath-13B) fine-tuned from LLaMA2-13B achieve slight improvements on English systems, but the performance results are lower than LLaMA2-13B on Chinese systems a lot,
which indicates that domain-specific fine-tuning may degrade the generalization abilities of LLMs.
(4). The mathematical models (i.e., InternLM2-Math-7B/20B and DeepSeekMath-7B) by continuing pretraining on the large-scale math-related dataset (>=100B tokens) show sufficient improvements when compared to models with similar size,
which indicates that large-scale pertaining is effective to improve the mathematical reasoning abilities.

\begin{figure}[t!]
    \centering
    \includegraphics[width=1.0\linewidth]{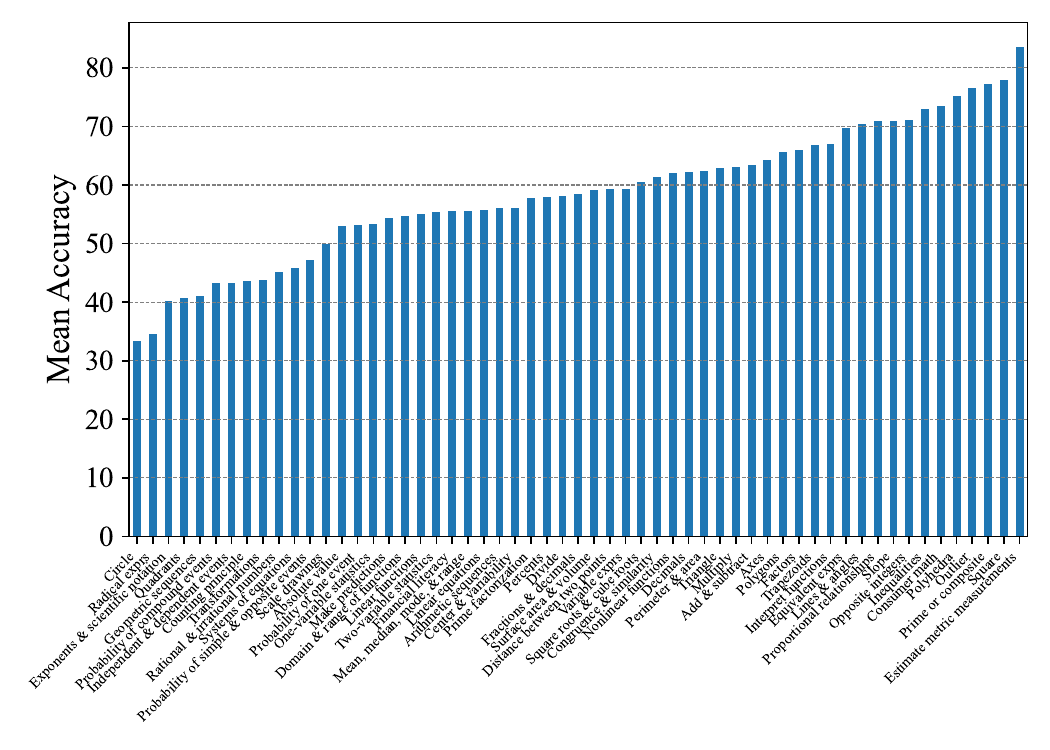}
    \vspace{-6mm}
    \caption{Mean concept accuracies on Middle-EN.} 
    \label{fig: mean_concept_acc_middle_en}
    \vspace{-3mm}
\end{figure}

\begin{figure}[t!]
    \centering
    \includegraphics[width=1.0\linewidth]{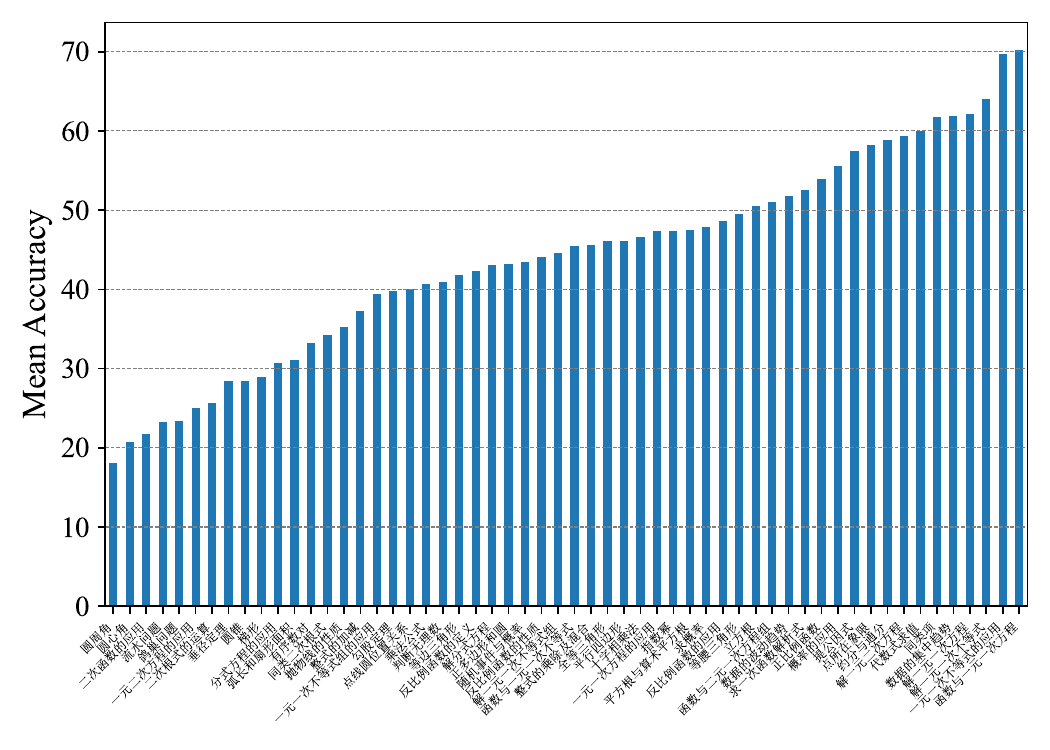}
    \vspace{-6mm}
    \caption{Mean concept accuracies on Middle-ZH.}
    \label{fig: mean_concept_acc_middle_zh}
    \vspace{-5mm}
\end{figure}

\paragraph{Average Concept-wised Accuracy.}
In Fig.~\ref{fig: mean_concept_acc_middle_en} and Fig.~\ref{fig: mean_concept_acc_middle_zh}, 
to better analyze the effectiveness of our \ourbench,
we further provide the concept-wised accuracies average on evaluated models for different mathematical concepts by zero-shot prompting on Middle-EN and Middle-ZH. (See Appendix~\ref{app: results} for more results on Elementary-EN and Elementary-ZH). 
In Fig.~\ref{fig: mean_concept_acc_middle_en} and Fig.~\ref{fig: mean_concept_acc_middle_zh},
we observe that 
the accuracies across concepts vary a lot for existing LLMs. For example, for Middle-ZH in Fig.~\ref{fig: mean_concept_acc_middle_zh}, around 18\% of concepts exhibit an accuracy lower than 30\%.
Thus,
to improve the mathematical abilities of LLMs,
these concepts with large room for improvement should be given the highest priority,
which further shows the advantage of ConceptMath.

\begin{figure}[t!]
    \centering
    \includegraphics[width=1.0\linewidth]{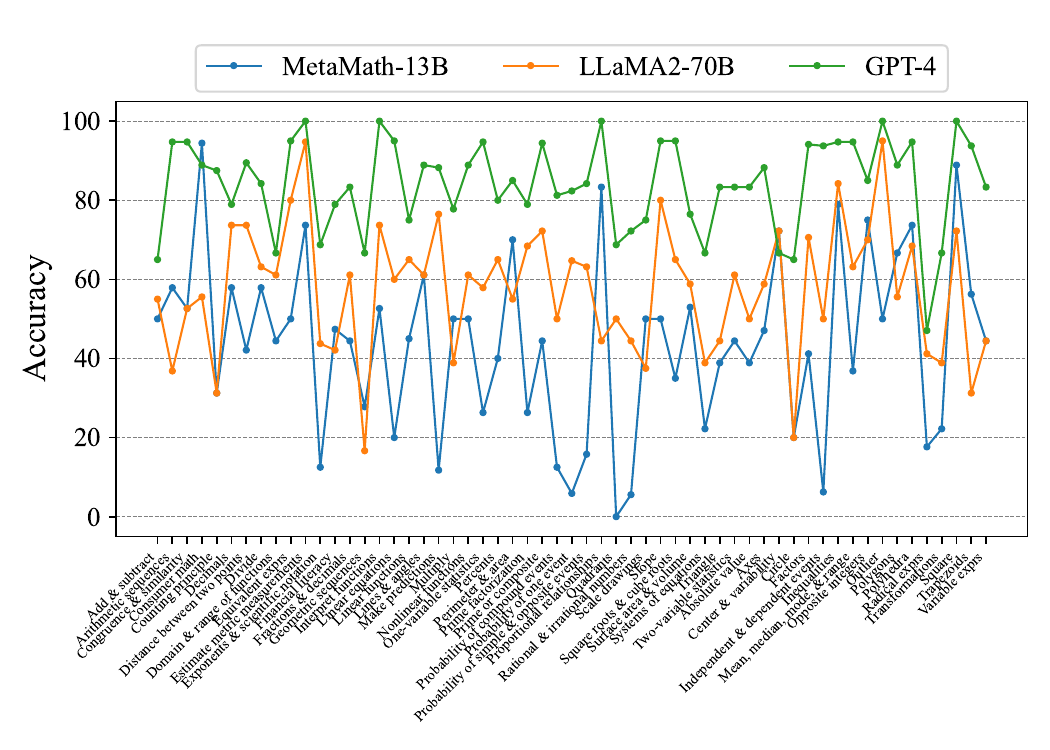}
        \vspace{-6mm}

    \caption{Concept accuracies on Middle-EN.}
    \label{fig: concept_acc_middle_en}
        \vspace{-5mm}

\end{figure}

\begin{figure}[t!]
    \centering
    \includegraphics[width=1.0\linewidth]{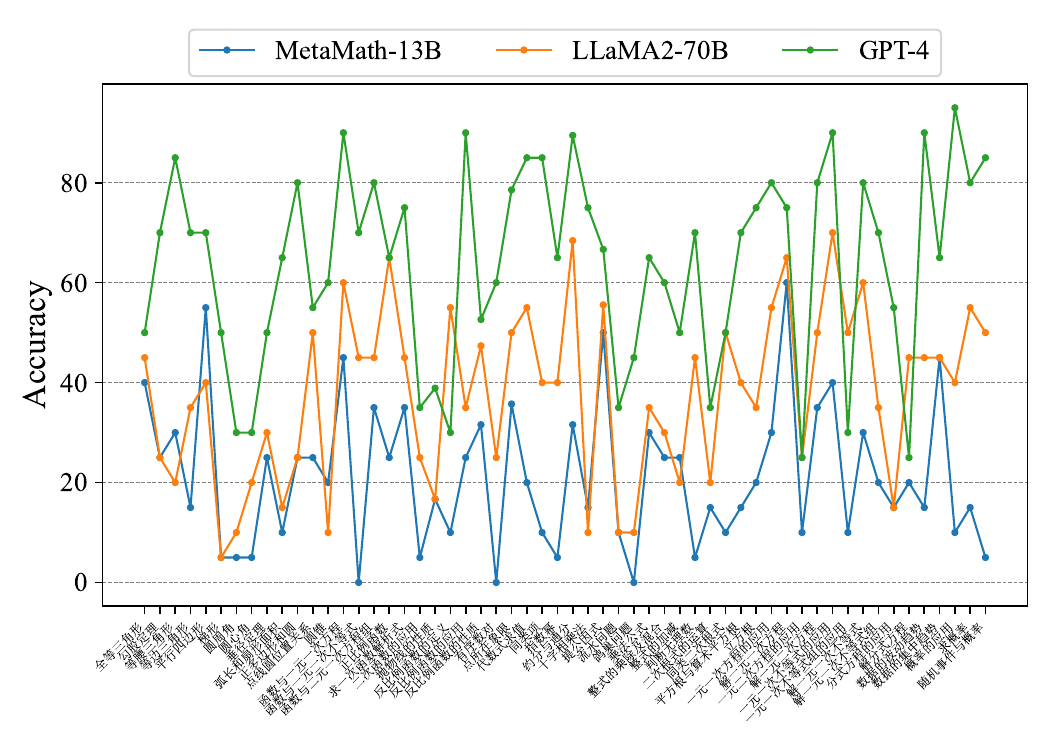}
    \vspace{-6mm}
    \caption{Concept accuracies on Middle-ZH.}
    \label{fig: concept_acc_middle_zh}
    \vspace{-4mm}
\end{figure}

\paragraph{Concept-wised Accuracy.}
Fig.~\ref{fig: concept_acc_middle_en} and Fig.~\ref{fig: concept_acc_middle_zh} show that most existing LLMs, whether open-sourced, closed-sourced, general-purpose, or math-specialized, exhibit notable differences in their concept accuracies in the zero-shot prompt setting. These disparities may stem from variations in training datasets, strategies, and model sizes, which suggests that apart from common weaknesses, each model possesses its unique areas of deficiency or shortcomings. For the sake of brevity in the presentation, we only show a subset of models on Middle-EN and Middle-ZH. The concept accuracies of Elementary-EN and Elementary-ZH systems and all results of all models can be found in Appendix~\ref{app: results}.

\begin{table*}[t]
\centering
\setlength\tabcolsep{11pt}
\small
\begin{tabular}{c|cccc|c}
\toprule
\multirow{1}{*}{\textbf{Model}} & \multicolumn{1}{c}{\textbf{Elementary-EN}} & \multicolumn{1}{c}{\textbf{Middle-EN}} & \multicolumn{1}{c}{\textbf{Elementary-ZH}} & \multicolumn{1}{c|}{\textbf{Middle-ZH}} & \multirow{1}{*}{\textbf{Avg. $\downarrow$}} \\ 
 \midrule
Yi-6B	                         	&5.30  / 1.73 	&5.21 / 1.37 	&0.04 / 0.20 	&0.36 / 0.35    &2.73 / 0.91 \\
ChatGLM3-6B	                     	&7.42  / 0.22 	&7.55 / 0.23 	&0.11 / 0.02 	&0.35 / 0.05    &3.86 / 0.13 \\
InternLM2-Math-7B	             	&7.42  / 0.22 	&7.55 / 0.23 	&0.11 / 0.02 	&0.35 / 0.05    &3.86 / 0.13 \\
InternLM2-7B	                 	&5.36  / 1.03 	&5.27 / 0.84 	&0.01 / 0.37 	&0.33 / 0.49    &2.74 / 0.68 \\
\midrule
MAmmoTH-13B	                     	&7.67  / 0.47 	&7.97 / 0.46 	&0.00 / 0.03 	&0.35 / 0.03    &4.00 / 0.25 \\
WizardMath-13B	                 	&8.41  / 0.35 	&8.23 / 0.34 	&0.00 / 0.02 	&0.55 / 0.02    &4.30 / 0.18 \\
MetaMath-13B	                 	&7.67  / 0.47 	&7.97 / 0.46 	&0.00 / 0.03 	&0.35 / 0.03    &4.00 / 0.25 \\
Baichuan2-13B	                 	&7.20  / 1.43 	&6.58 / 1.18 	&0.05 / 0.54 	&0.41 / 0.65    &3.56 / 0.95 \\
LLaMA2-13B	                     	&6.80  / 0.73 	&6.36 / 0.64 	&0.01 / 0.15 	&0.56 / 0.16    &3.43 / 0.42 \\
Qwen-14B	                     	&11.04 / 1.58 	&9.73 / 1.08 	&1.43 / 1.27 	&0.70 / 0.93    &5.73 / 1.22 \\
\midrule
InternLM2-Math-20B	             	&5.58  / 1.30 	&5.51 / 0.99 	&0.03 / 0.47 	&0.34 / 0.47    &2.86 / 0.81 \\
InternLM2-20B	                 	&7.20  / 1.43 	&6.58 / 1.18 	&0.05 / 0.54 	&0.41 / 0.65    &3.56 / 0.95 \\
\midrule
GPT-3.5                             &9.48 / -      &9.21 / -      &0.00 / - 	    &0.31 / -       &4.75 / -    \\ 
GPT-4                              & 8.68 / -      &8.24 / -      &0.15 / - 	    &0.68 / -       &4.44 / -    \\
\bottomrule
\end{tabular}
\vspace{-1mm}
\caption{Data contamination rate of LLMs. We provide two different contamination detection methods. The values in the table represent ``Rouge / Prob''. Note that the second method based on output probability distributions can only be applied to the open-source models.}
\label{tab:contami}
\vspace{-2mm}
\end{table*}
\subsection{Analysis}

\paragraph{Contamination.}
To determine whether a text  is in the pretraining data of
a LLM,
we provide two different contamination detection methods (i.e., Rouge-based and Prob-based methods) to analyze our  ConceptMath in Table~\ref{tab:contami}.
Specifically,
for the Rouge-based method,
we just input the first 50\% of the question as the input and compute the Rouge-L score between the generation results and the ground-truth label of the last 50\% of the text,
where a lower Rouge-L score means a lower contamination rate.
For the Prob-based method,
we follow~\cite{shi2023detecting} to use the  MIN-K\% probability metric, which first gets the probability for each token in the test, and selects the K\%
tokens with minimum probabilities and calculates their average log-likelihood. If the average log-likelihood is high, the text is likely in the pretraining data. Note that we choose $K$ as 10 in our setting.
In Table~\ref{tab:contami},
we observe that the contaminate rates on our ConceptMath are very low,
which means that our ConceptMath can provide a reasonable evaluation for existing LLMs.

\begin{table*}[t]
\centering
\small
\setlength\tabcolsep{10pt}
\resizebox{0.99\linewidth}{!}{
\begin{tabular}{ccc|l}
\toprule
\textbf{Concept} & \textbf{Concept Acc} & \textbf{Error Type} & \textbf{Example} \\
\midrule
\multirow{6}{*}{Perimeter} &
  \multirow{6}{*}{26.32} &
  Calculation Error &
  \begin{tabular}[c]{@{}l@{}}
\textcolor{blue}{Q}: Find the perimeter of a 16 by 18 feet rectangle.\\ \textcolor{blue}{A}: The perimeter is \textcolor{red}{16 + 18 + 16 + 18 = 72} feet.
\end{tabular}\\ \cmidrule{3-4}
 &
&
  Knowledge Error &
  \begin{tabular}[c]{@{}l@{}}
\textcolor{blue}{Q}: The perimeter of a rectangular sticker is 14 centimeters. \\The sticker is 2 centimeters wide. How tall is it?\\ \textcolor{blue}{A}: \textcolor{red}{The height equals the perimeter subtracts the width}.\\The height of the rectangle is $14-12=12$ centimeters.
\end{tabular}\\ \midrule
Polygons &
  11.11 &
  Knowledge Error &
  \begin{tabular}[c]{@{}l@{}}
\textcolor{blue}{Q}: A regular polygon has an interior angle that measures \\120 degrees. How many sides does this polygon have?\\\textcolor{blue}{A}: Let $n$ be the number of sides of the polygon. We have \\that \textcolor{red}{$n \times 120 = 360$}. Hence: $n = 3$.
\end{tabular}\\ 
\midrule
Cylinders &
  11.76 &
  Calculation Error &
  \begin{tabular}[c]{@{}l@{}}
\textcolor{blue}{Q}: How much soup can a cylinder-shaped can store, if the \\radius of the can is 3 inches and the height is 5 inches?
\\\textcolor{blue}{A}: ... The volume of the can is V = \textcolor{red}{$\pi\times3^2\times5 = 30\pi$}.
\end{tabular}\\
\bottomrule
\end{tabular}
}
\vspace{-1mm}
\caption{Error reasons of several unmastered concepts of LLaMA2-13B.}
\vspace{-1mm}
\label{tab: samples}
\end{table*}

\paragraph{Unmastered Concepts.}

We also highlight the several unmastered concepts of the LLaMA2-13B in Table~\ref{tab: samples}, 
which shows ConceptMath is effective in guiding further refinement of existing LLMs.

\begin{table}[ht]
\resizebox{0.99\linewidth}{!}{
\begin{tabular}{c|ccc}
\toprule
\multirow{3}{*}{\textbf{Models}} &
\multirow{3}{*}{\textbf{LLaMA2}} &
\multirow{2}{*}{\textbf{LLaMA2}} & 
\textbf{LLaMA2} \\
& & \multirow{2}{*}{\textbf{(w/ MMQA)}} &\textbf{(w/ MMQA}\\
& & & \textbf{\&CS)} \\
\midrule
Cones &0.00 & 17.65 & 23.53\\ 
Spheres & 5.88 & 29.41 & 35.29 \\ 
Polygons & 11.11 & 61.11 & 66.67 \\ 
Rational Number & 11.76 & 23.53 & 52.94 \\ 
Cylinders & 11.76 & 35.29 & 47.06 \\ 
Angles & 11.76 & 47.06 & 58.82  \\
Probability & 18.75 & 25.00 & 75.00 \\ 
Perimeter& 26.32 & 42.11 & 63.16 \\ 
Volume & 27.78 & 38.89 & 66.67 \\ 
Proportional & 27.78 & 33.33 & 44.44 \\ \midrule
\multirow{1}{*}{\textbf{Avg Acc.}}
& \multirow{2}{*}{15.29} & \multirow{2}{*}{36.88} & \multirow{2}{*}{\textbf{53.36}} \\
(over 10 concepts) & & & \\
\midrule
\multirow{1}{*}{\textbf{Avg Acc.}}
& \multirow{2}{*}{51.94} & \multirow{2}{*}{58.14} & \multirow{2}{*}{\textbf{60.67}} \\
(over 33 concepts) & & & \\
\midrule
\textbf{Overall Acc.} & 44.02 & 53.94 & \textbf{59.29} \\ 
\bottomrule
\end{tabular}}
\caption{Results of fine-tuning models. ``MMQA'' and ``CS'' denote MetaMathQA and our constructed Concept-Specific training datasets, respectively. Introducing CS data specifically for the bottom 10 concepts significantly enhances these concepts' performance, while slightly improving the performance across the remaining 33 concepts.}
\label{tab:concept_enhance}
\vspace{-2mm}
\end{table}

\paragraph{Evaluation Prompting.} Different from the few-shot or cot prompting evaluation that can boost closed-source models, we find that zero-shot prompting is more effective for certain open-source LLMs in Table~\ref{tab:mainresults}. This disparity may arise either because the models are not sufficiently powerful to own mathematical CoT capabilities~\cite{yu2023metamath, wei2022chain} or because these models have already incorporated CoT data during training~\cite{flan}. Consequently, to ensure a comprehensive analysis, we have employed all three prompting methods for evaluation.

\paragraph{Efficient Fine-tuning.}
To show the effect of efficient fine-tuning,
we take the LLaMA2-13B as an example in Table~\ref{tab:concept_enhance}.
Specifically,
for LLaMA2-13B,
we first select 10 concepts with the lowest accuracies in Elementary-EN.
Then, we crawl 495 samples (about 50 samples per concept) using the trained classifier as the Concept-Specific (\textbf{CS}) training data (See Appendix~\ref{app: eft} for more details).
Meanwhile,
to avoid overfitting,
we introduce the MetaMathQA (\textbf{MMQA}~\cite{yu2023metamath} ) data to preserve general mathematical abilities.
After that,
we can fine-tune LLaMA2-13B by only using MMQA (i.e., LLaMA2 (w/ MMQA)),
or using both  MMQA and CS data  (i.e., LLaMA2 (w/ MMQA \& CS)).
In Table~\ref{tab:concept_enhance},
we observe that LLaMA2 (w/ MMQA \& CS) archives 
significant improvements on the lowest 10 concepts and preserves well on the other  33 concepts,
which shows the effect of efficient fine-tuning and the advantages of our ConceptMath.

\section{Related Work}
\paragraph{Large Language Models for Mathematics.}
Large Language Models (LLMs) such as GPT-3.5 and GPT-4 have exhibited promising capabilities in complex mathematical tasks.
However, the proficiency of open-source alternatives like LLaMA~\citep{touvron2023llama} and LLaMA2~\citep{touvron2023llama2} remains notably inferior on these datasets, particularly in handling non-English problems. In contrast, models like Baichuan2~\citep{baichuan} and Qwen~\citep{qwen} pretrained on multilingual datasets (i.e., Chinese and English) have achieved remarkable performance. 
Recently, 
many domain-specialized math language models have been proposed. For example, MetaMath~\citep{yu2023metamath} leverages the LLaMA2 models and finetunes on the constructed MetaMathQA dataset.
MAmmoTH~\citep{yue2023mammoth} synergizes Chain-of-Thought (CoT) and Program-of-Thought (PoT) rationales. 
\paragraph{Mathmatical Reasoning Benchmarks.}
Recently, many mathematical datasets~\cite{roy2015solving,koncel2015parsing,lu2022dynamic,huang2016well,miao2020diverse,patel2021nlp} have been proposed.
For example,
SingleOp \citep{roy2015reasoning}, expands the scope to include more complex operations like multiplication and division.
Math23k \citep{wang2017deep} gathers 23,161 problems labeled with structured equations and corresponding answers.
GSM8K~\citep{cobbe2021training} is a widely used dataset,
which requires a sequence of elementary calculations with basic arithmetic operations.
\paragraph{Fine-Grained Benchmarks.}
Traditional benchmarks focus on assessing certain abilities of models on one task~\cite{guo2023owl,wang2023rolellm,liu2020block,guo2022lvp,chai2024xcot,liu20242,guo2024logformer,guo2023adaptive,bai2023griprank,liu2022cross,guo2023m2c,mtbench,liu2021dam}
(e.g., reading comprehension~\citep{rajpurkar2018know}, machine translation~\citep{bojar-etal-2014-findings}, and summarization~\citep{narayan-etal-2018-dont}).
For example, the GLUE benchmark~\citep{wangglue} combines a collection of tasks, and has witnessed superhuman model performance for pretraining models~\citep{kenton2019bert,radford2019language}
\citep{mmlu} introduced MMLU, a benchmark with multiple-choice questions across 57 subjects including STEM, humanities, and social sciences, for assessing performance and identifying weaknesses. \citep{srivastava2022beyond} proposed BIG-bench with over 200 tasks.
To enhance the mathematical capabilities of LLMs, we introduce a comprehensive mathematical reasoning ConceptMath dataset designed to assess model performance across over 200 diverse mathematical concepts in both Chinese and English.
\vspace{1mm}
\section{Conclusion}
\vspace{1mm}
We introduce a new bilingual concept-wise math reasoning dataset called ConceptMath to assess models across a diverse set of concepts. 
First,  ConceptMath covers more than 200 concepts across elementary and middle schools for mainstream English and Chinese systems. 
Second, we extensively evaluate existing LLMs by three prompting methods, which can guide further improvements for these LLMs on mathematical abilities.
Third,
we analyze the contamination rates, error cases and provide a simple and efficient fine-tuning strategy to enhance the weaknesses.
\paragraph{Limitations.}
Human efforts are required to carefully design the hierarchical systems of mathematical concepts. In the future,
we have three plans as follows:
(1) Extend the input modality to multi-modalities.
(2) Extend the education systems to high school and college levels.
(3) Extend the reasoning abilities to more STEM fields.



\bibliography{custom}
\bibliographystyle{acl_natbib}

\appendix
\clearpage
\section{Details on the ConceptMath}
\label{app: concept}
\begin{figure*}[!ht]
    \centering
    \includegraphics[width=1.0\linewidth]{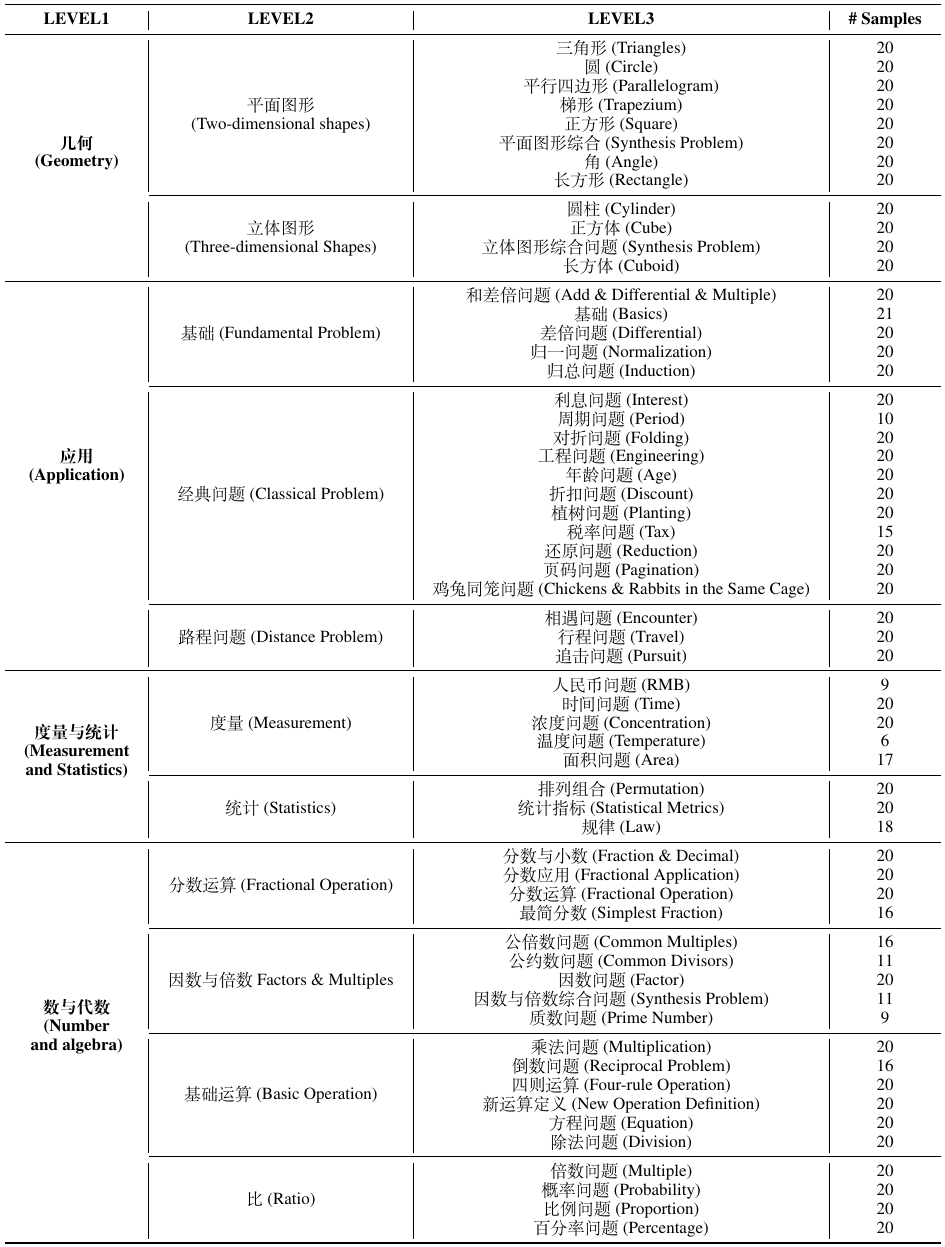}
    \caption{Details of the hierarchical concepts in Elementary-ZH.}
    \label{tab: ele-zh}
\end{figure*}

As shown in Table~\ref{tab: ele-en}, Table~\ref{tab: mid-en}, Table~\ref{tab: ele-zh} and Table~\ref{tab: mid-zh},
we have provided the details on the three-level hierarchical system of our ConceptMath for better illustration.

\begin{figure*}[h]
    \centering
    \includegraphics[width=1.0\linewidth]{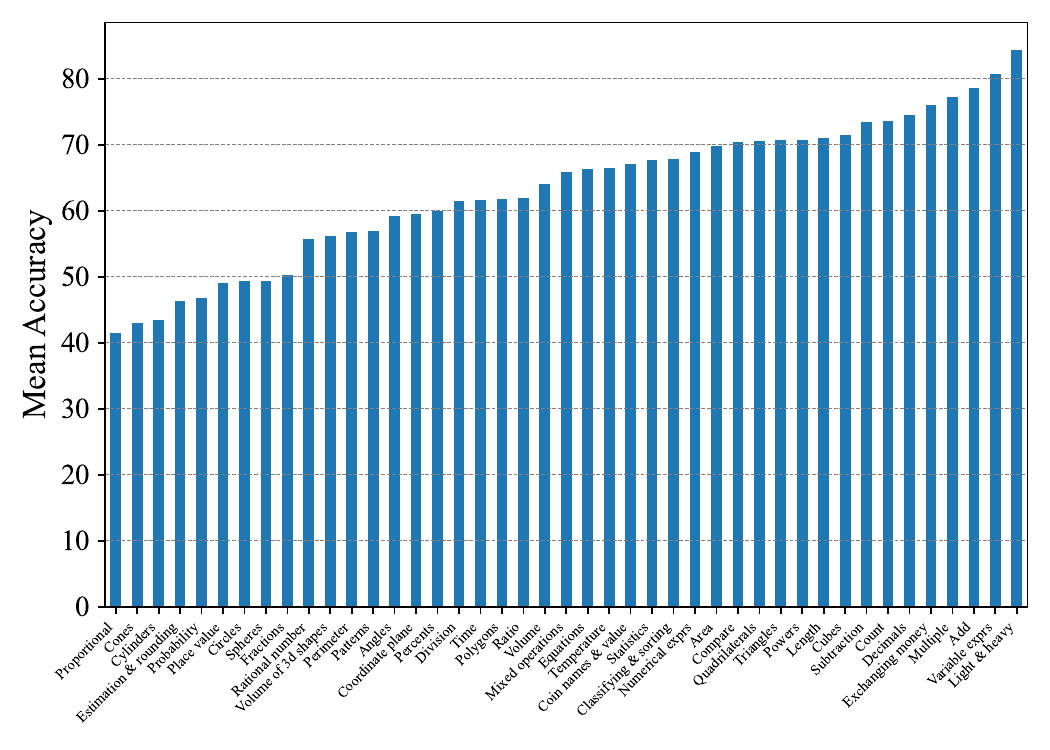}
    \caption{Mean concept accuracies of Elementary-EN.}
    \label{fig: mean_concept_acc_element_en}
\end{figure*}

\begin{figure*}[h]
    \centering
    \includegraphics[width=1.0\linewidth]{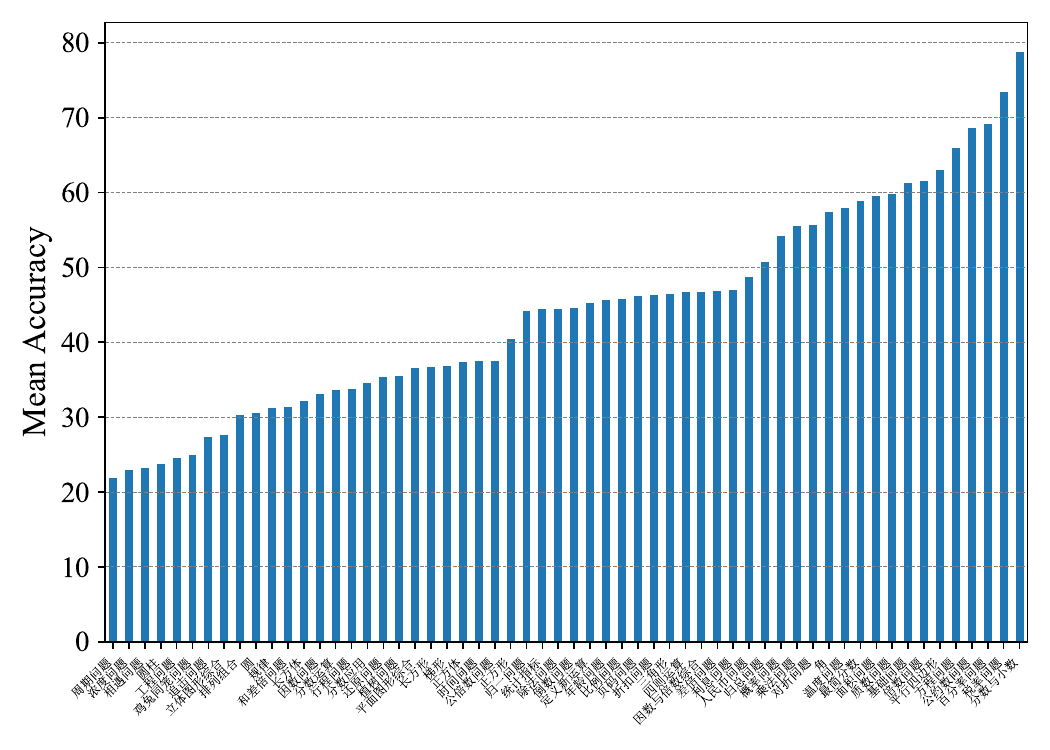}
    \caption{Mean concept accuracies of Elementary-ZH.}
    \label{fig: mean_concept_acc_element_zh}
\end{figure*}

\begin{figure*}[h]
    \centering
    \includegraphics[width=1.0\linewidth]{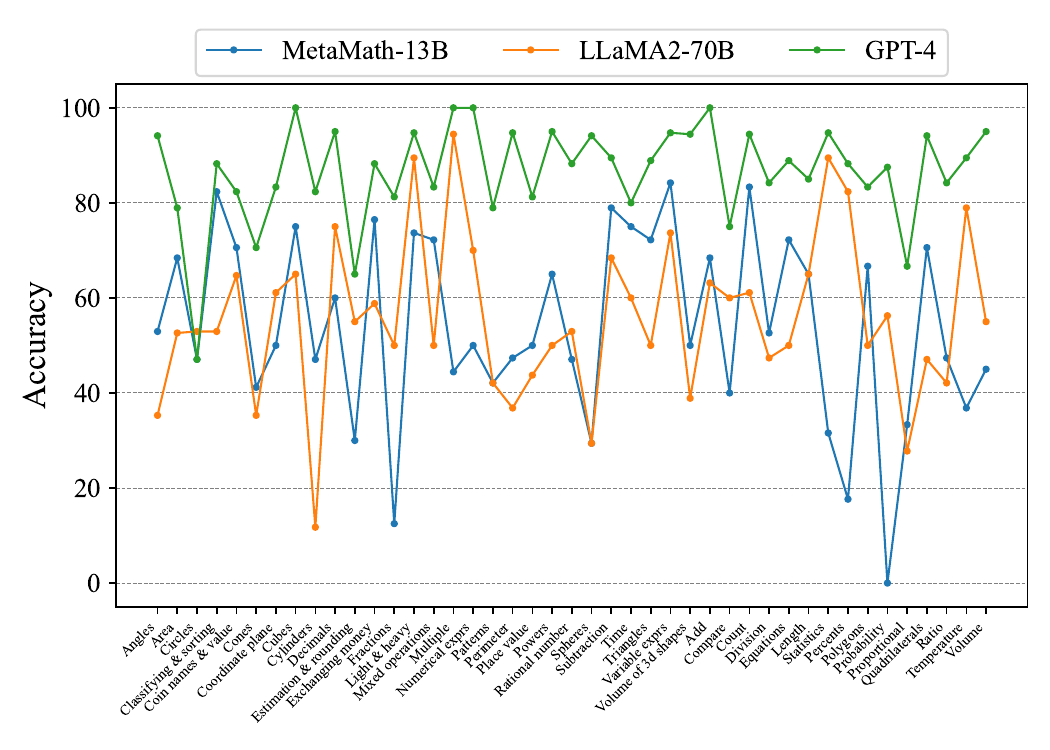}
    \caption{Concept accuracies on Elementary-EN.}
    \label{fig: concept_acc_element_en}
\end{figure*}

\begin{figure*}[h]
    \centering
    \includegraphics[width=1.0\linewidth]{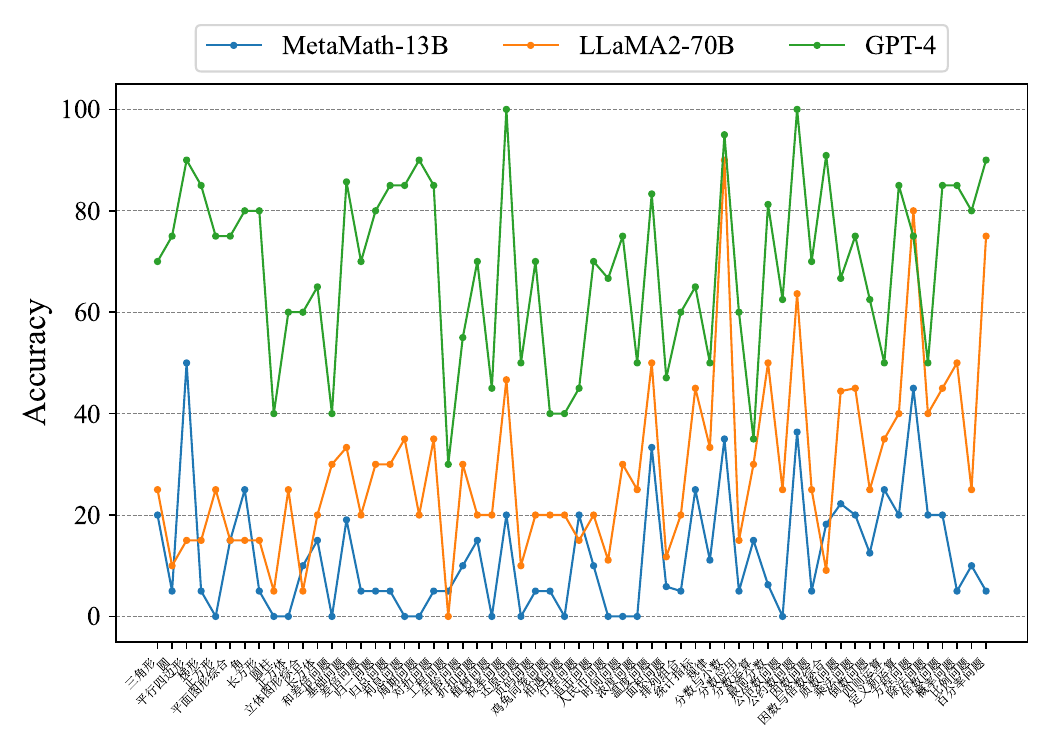}
    \caption{Concept accuracies on Elementary-ZH.}
    \label{fig: concept_acc_element_zh}
\end{figure*}

\begin{figure*}[h]
  \centering
\begin{subfigure}[b]{0.49\textwidth}
\includegraphics[width=1.0\linewidth]{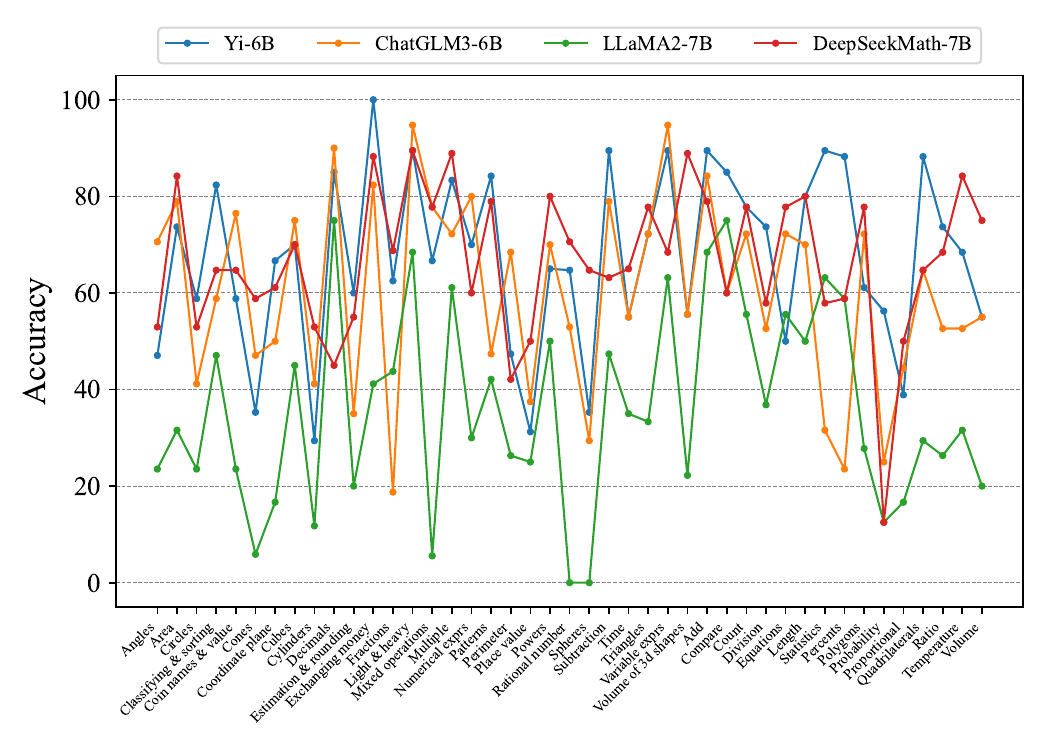}
\end{subfigure}
\begin{subfigure}[b]{0.49\textwidth}
    \includegraphics[width=1.0\linewidth]{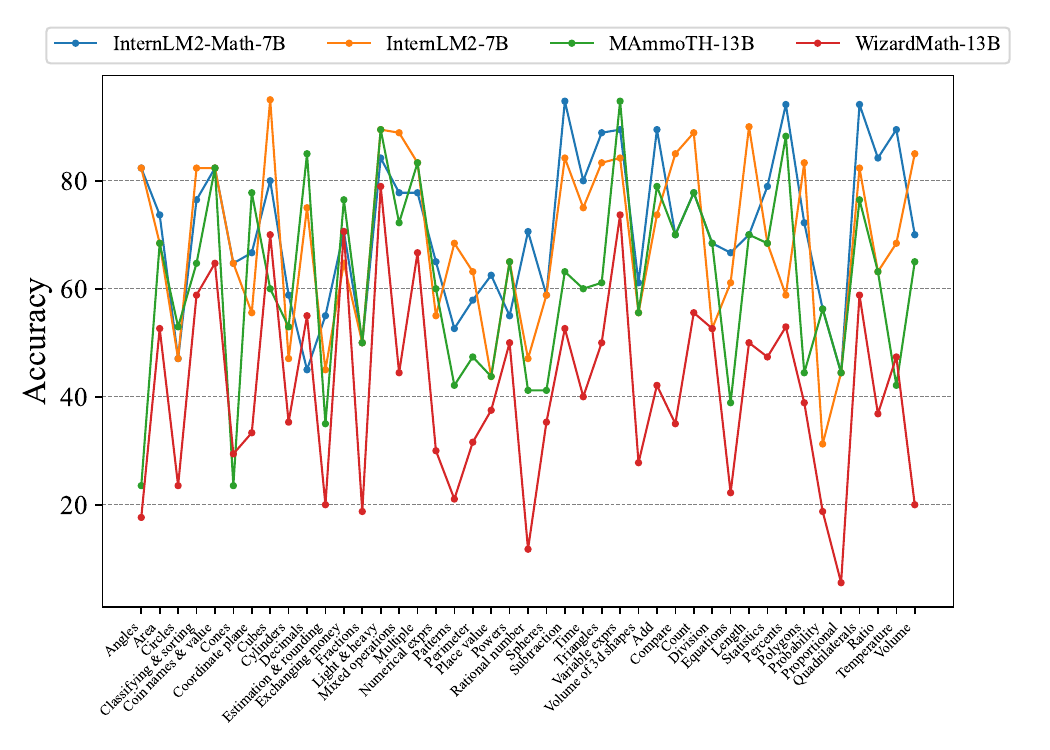}
\end{subfigure}
\begin{subfigure}[b]{0.49\textwidth}
    \includegraphics[width=1.0\linewidth]{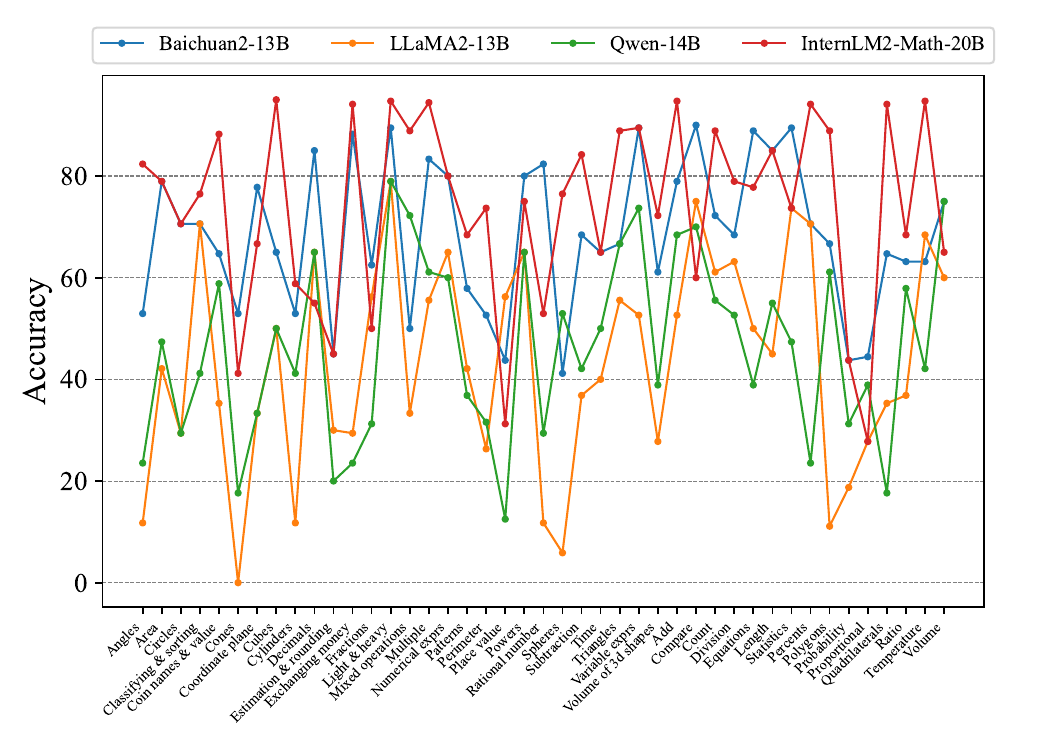}
\end{subfigure}
\begin{subfigure}[b]{0.47\textwidth}
    \includegraphics[width=0.98\linewidth]
    {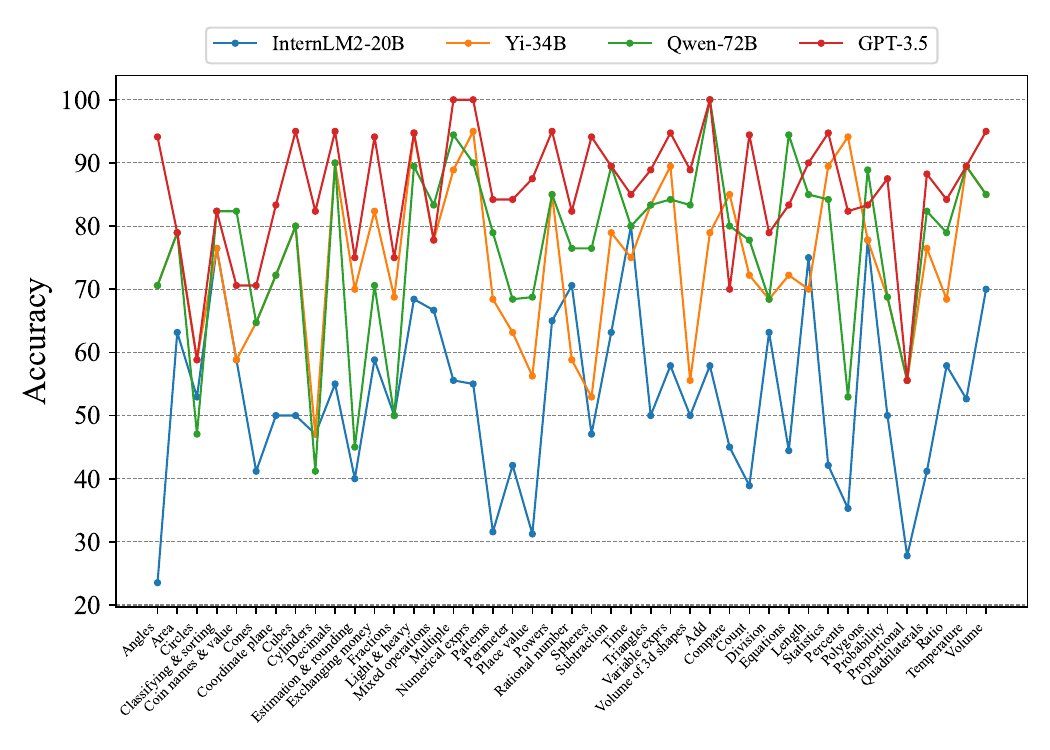}
\end{subfigure}
  \caption{Concept accuracies on Elementary-EN of more models.}
 \label{fig: all_model_concept_acc_element_en}
 \vspace{-2mm}
\end{figure*}

\begin{figure*}[h]
  \centering
\begin{subfigure}[b]{0.49\textwidth}
\includegraphics[width=1.0\linewidth]{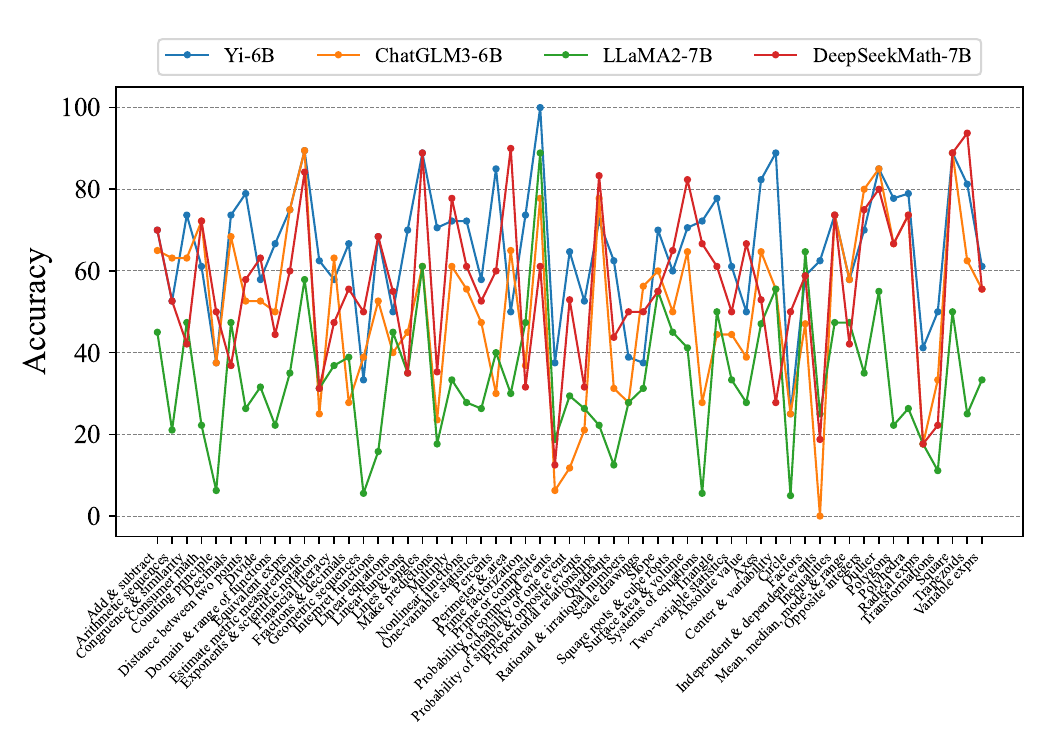}
\end{subfigure}
\begin{subfigure}[b]{0.49\textwidth}
    \includegraphics[width=1.0\linewidth]{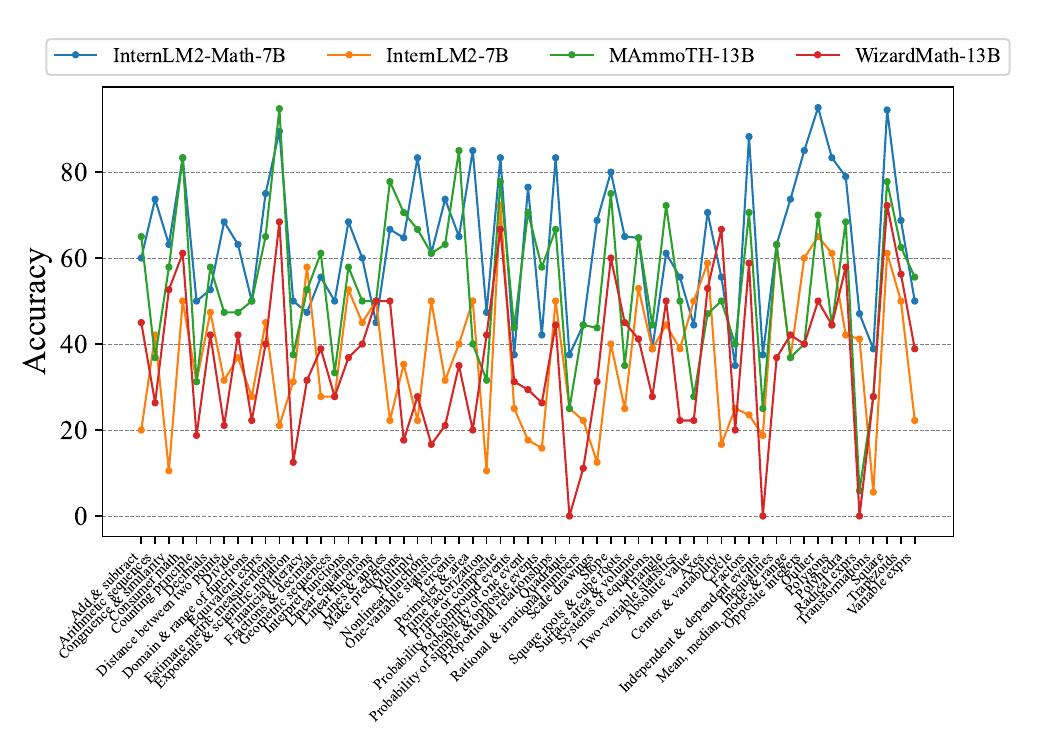}
\end{subfigure}
\begin{subfigure}[b]{0.49\textwidth}
    \includegraphics[width=1.0\linewidth]{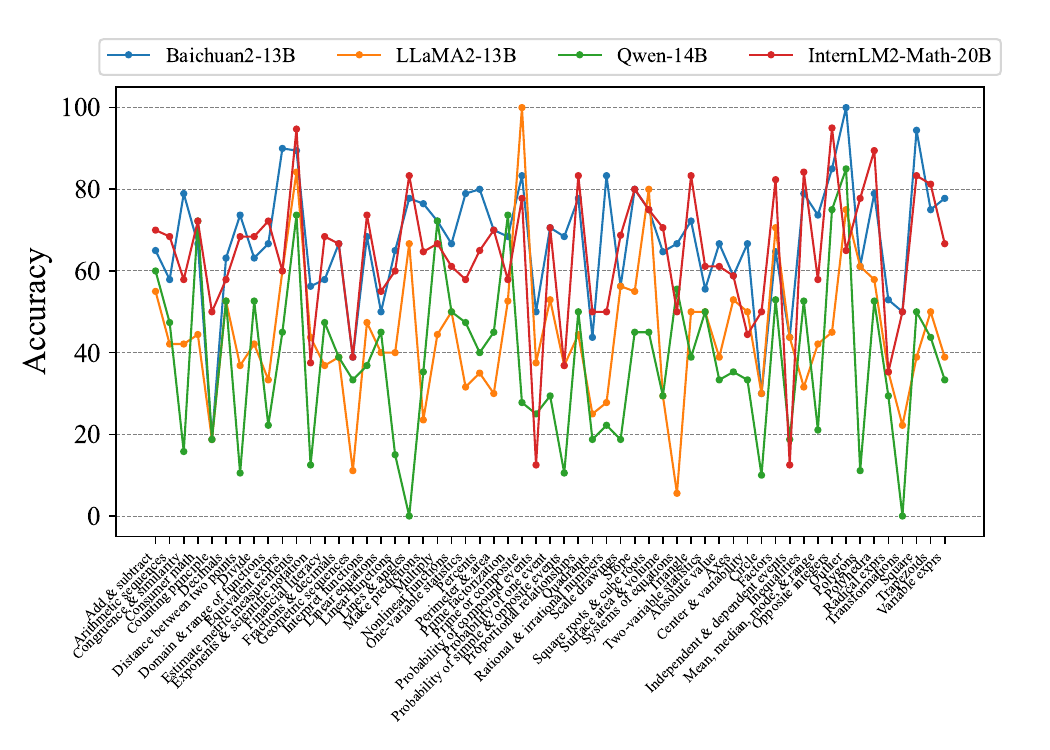}
\end{subfigure}
\begin{subfigure}[b]{0.49\textwidth}
    \includegraphics[width=0.98\linewidth]
    {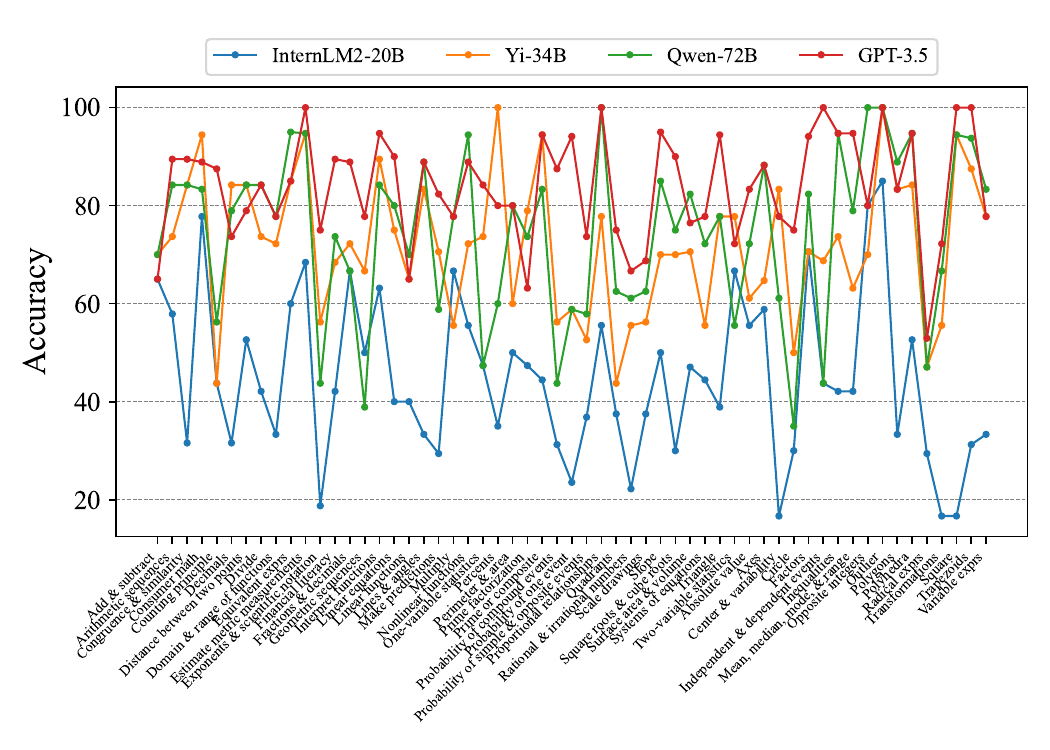}
\end{subfigure}
  \caption{Concept accuracies on Middle-EN of more models.}
 \label{fig: all_model_concept_acc_middle_en}
 \vspace{-2mm}
\end{figure*}

\begin{figure*}[h]
  \centering
\begin{subfigure}[b]{0.49\textwidth}
\includegraphics[width=1.0\linewidth]{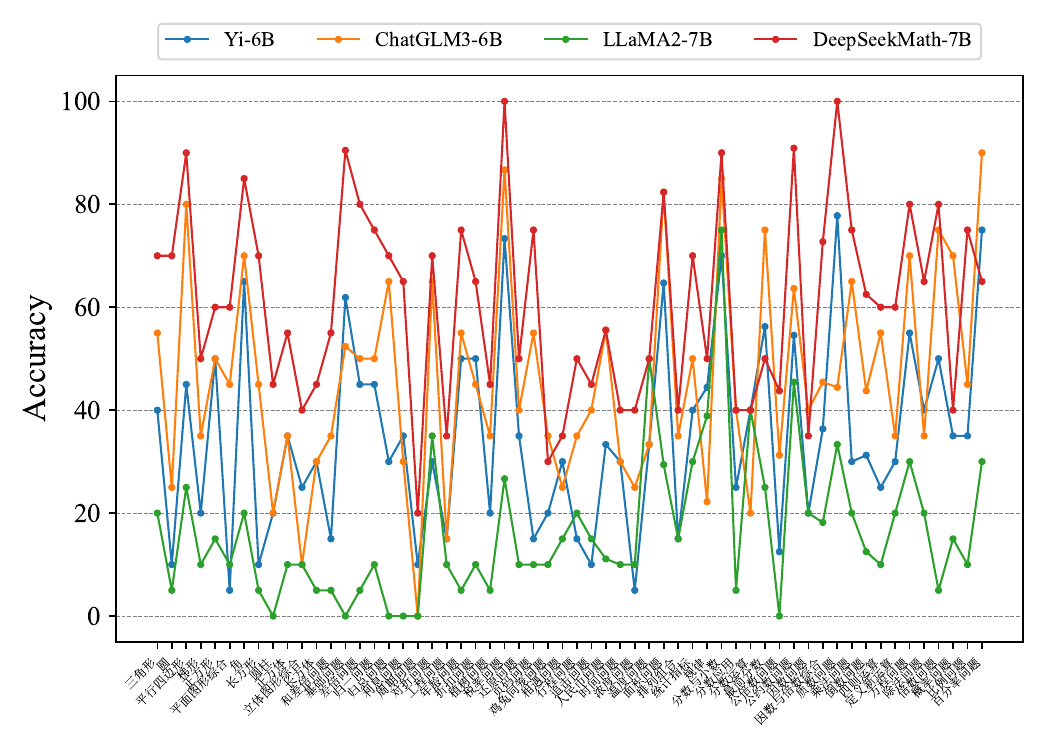}
\end{subfigure}
\begin{subfigure}[b]{0.49\textwidth}
    \includegraphics[width=1.0\linewidth]{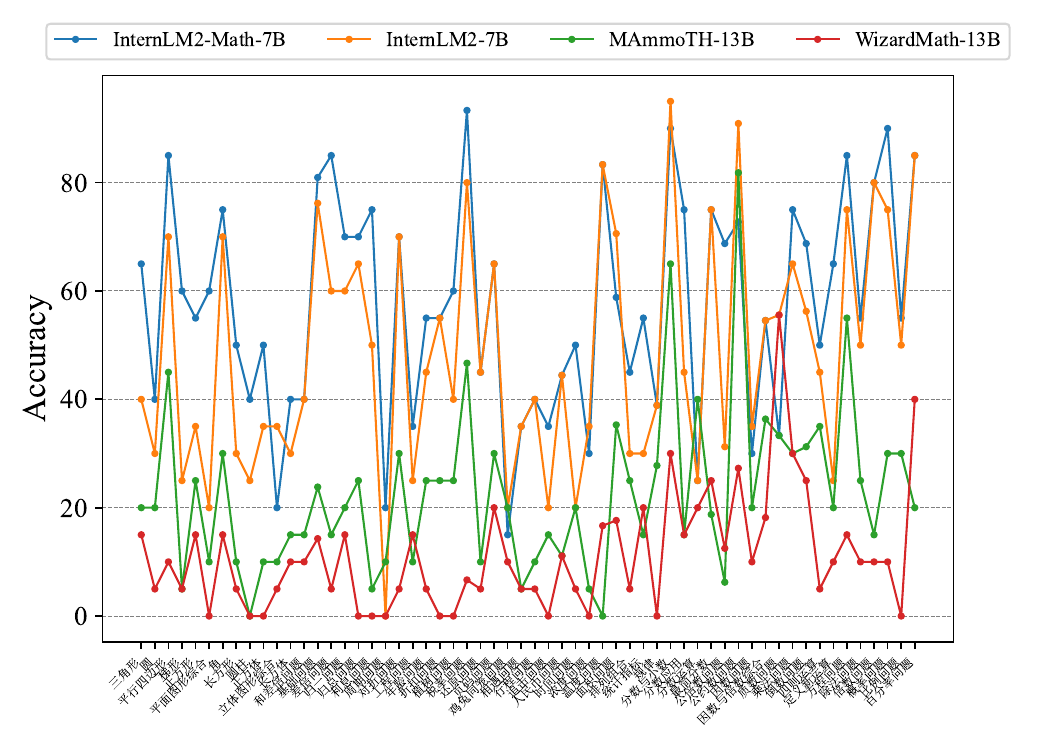}
\end{subfigure}
\begin{subfigure}[b]{0.49\textwidth}
    \includegraphics[width=1.0\linewidth]{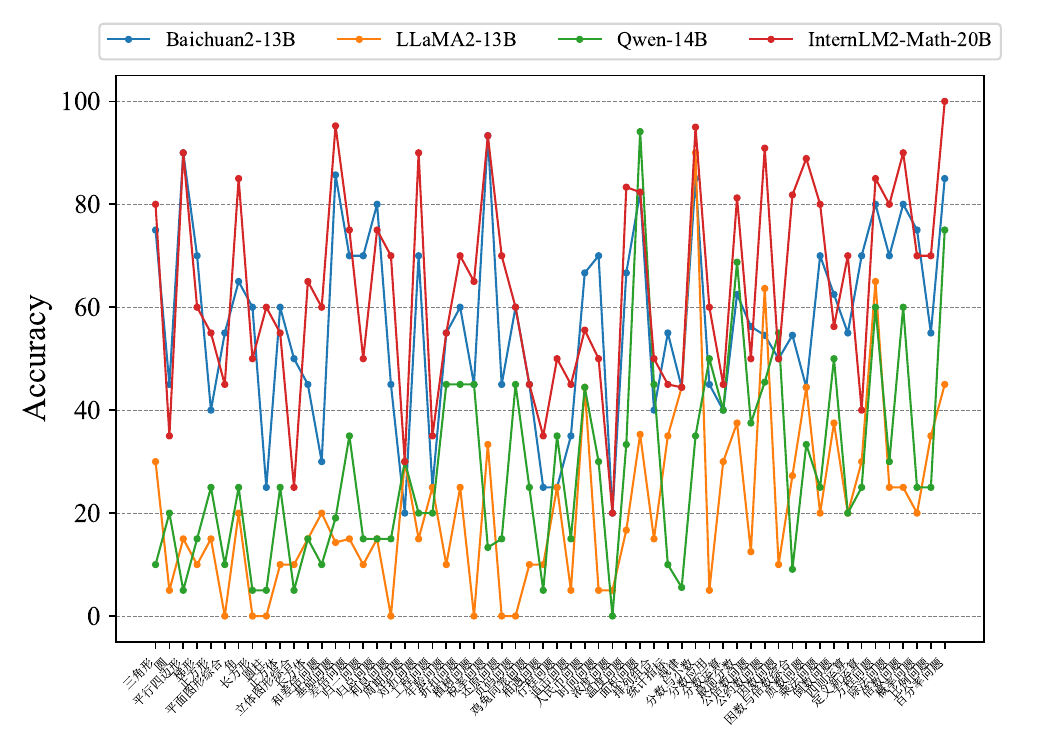}
\end{subfigure}
\begin{subfigure}[b]{0.49\textwidth}
    \includegraphics[width=0.95\linewidth]
    {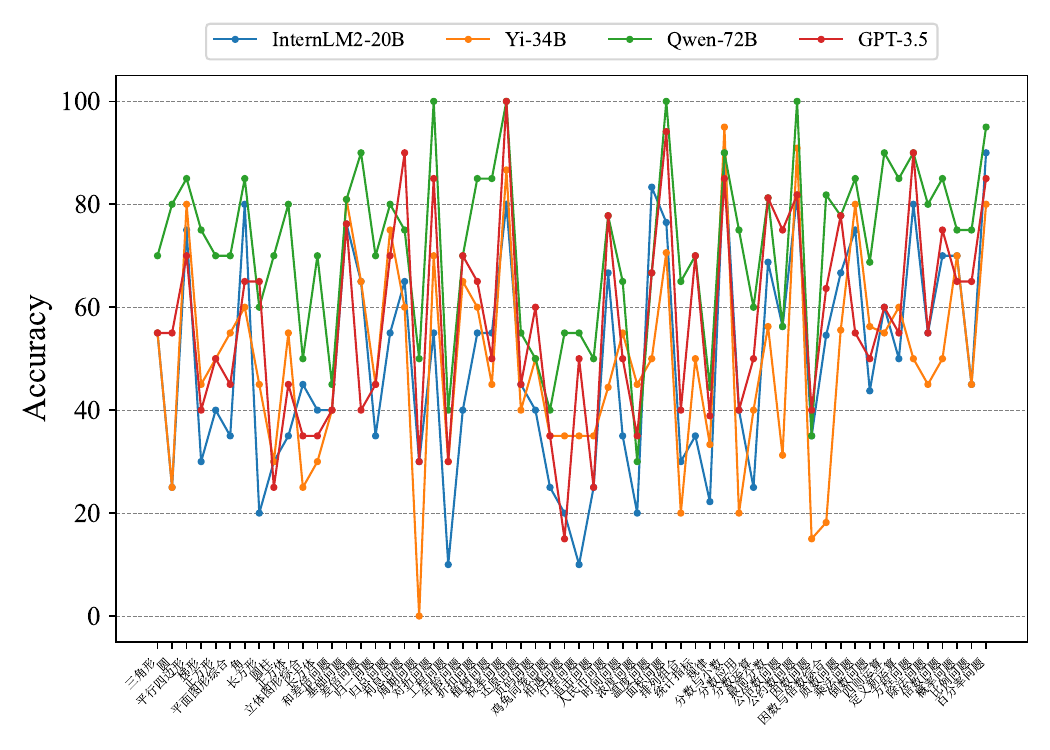}
\end{subfigure}
  \caption{Concept accuracies on Elementary-ZH of more models.}
 \label{fig: all_model_concept_acc_element_zh}
 \vspace{-2mm}
\end{figure*}

\begin{figure*}[h]
  \centering
\begin{subfigure}[b]{0.49\textwidth}
\includegraphics[width=1.0\linewidth]{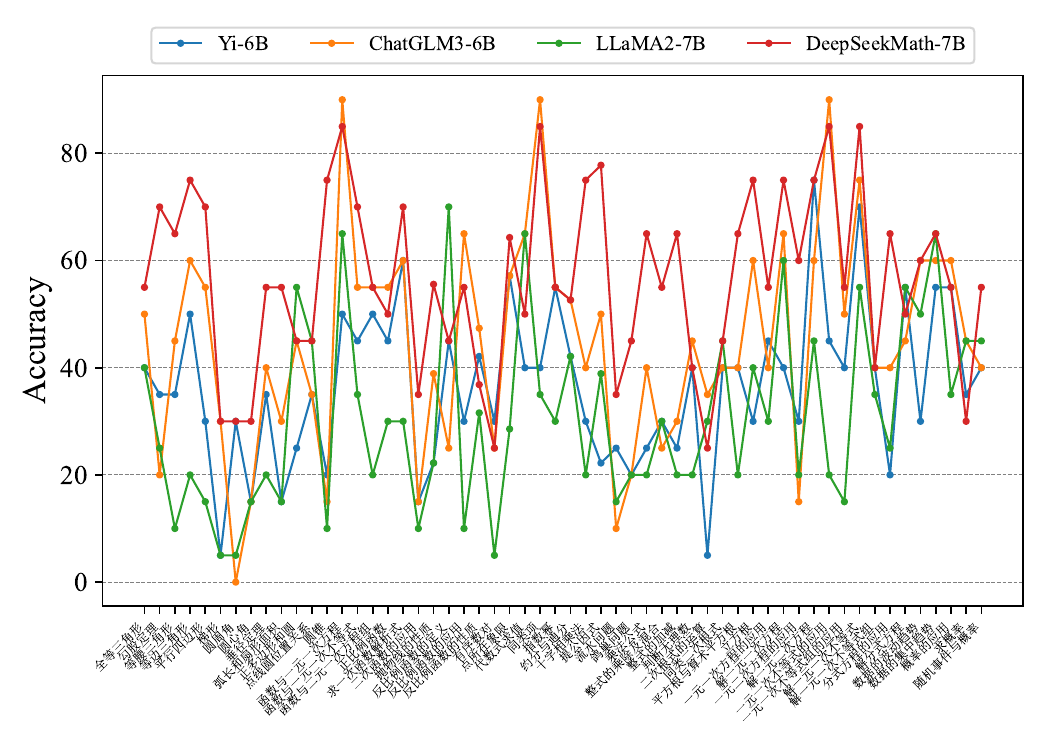}
\end{subfigure}
\begin{subfigure}[b]{0.49\textwidth}
    \includegraphics[width=1.0\linewidth]{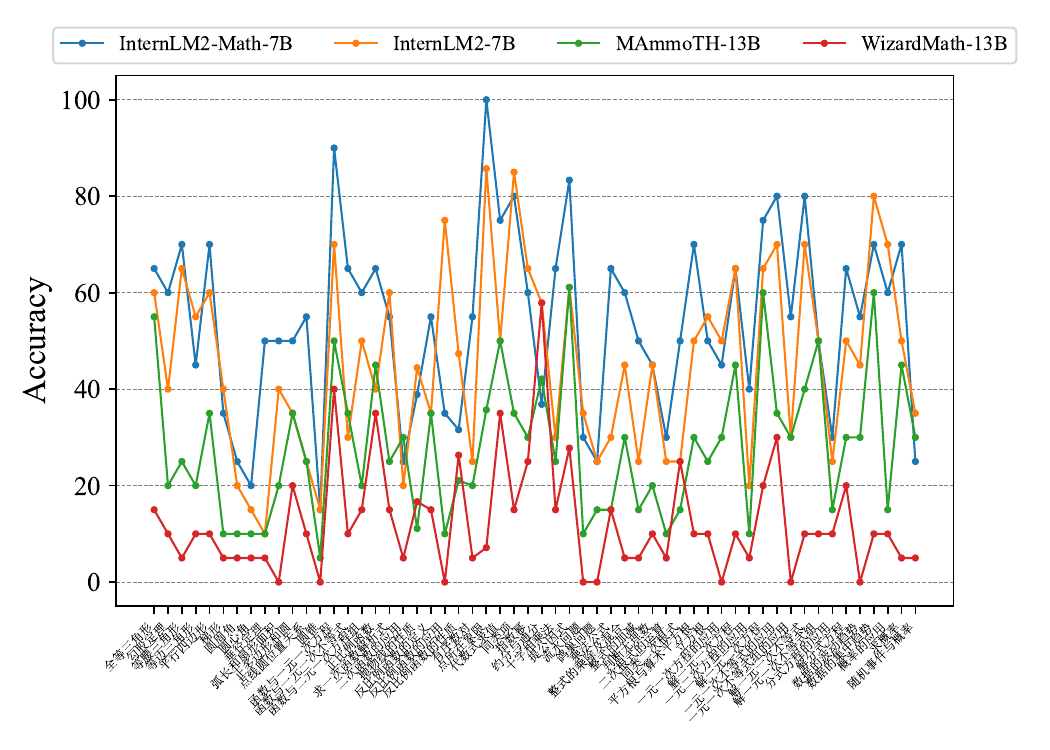}
\end{subfigure}
\begin{subfigure}[b]{0.49\textwidth}
    \includegraphics[width=1.0\linewidth]{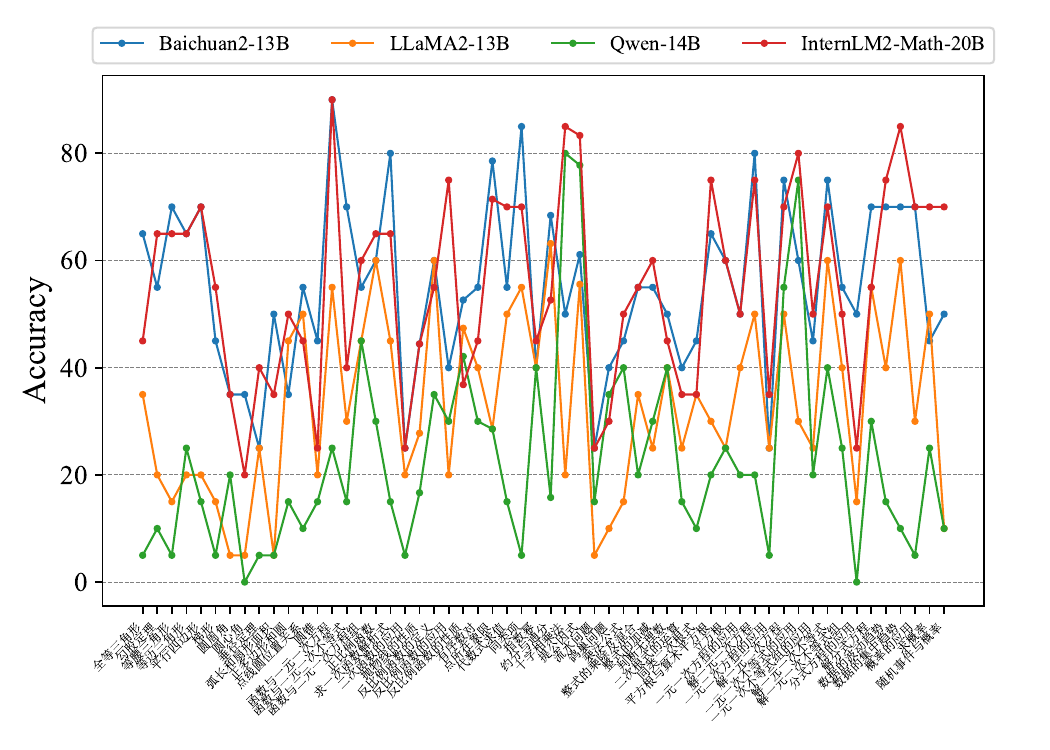}
\end{subfigure}
\begin{subfigure}[b]{0.49\textwidth}
    \includegraphics[width=0.98\linewidth]
    {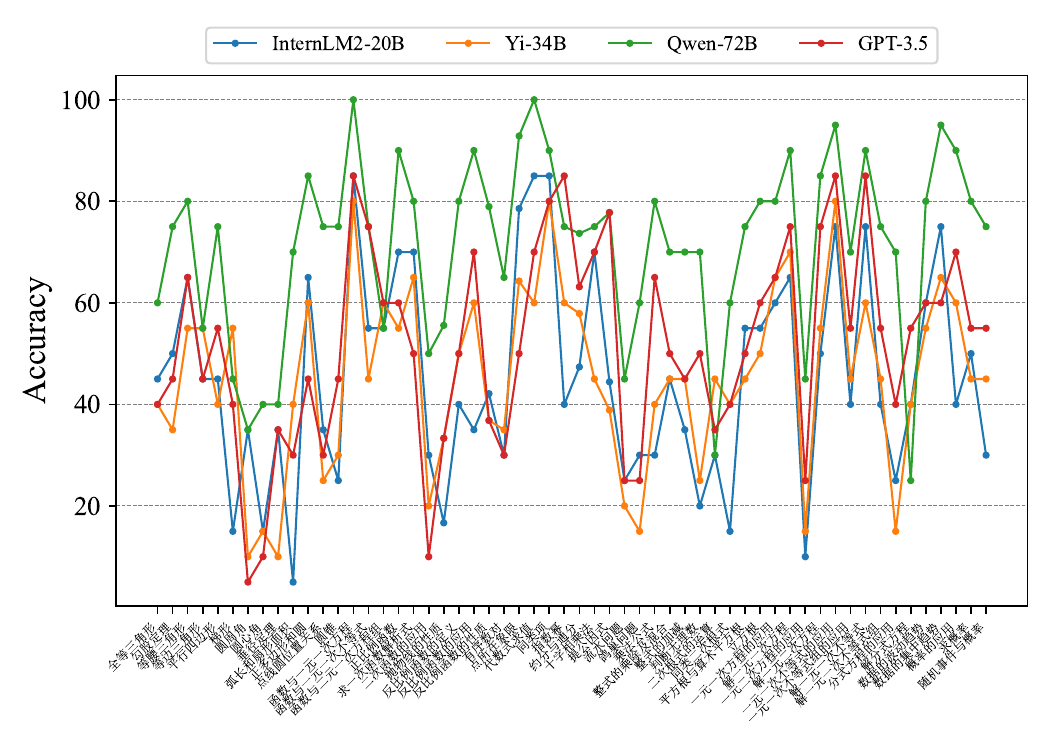}
\end{subfigure}
  \caption{Concept accuracies on Middle-ZH of more models.}
 \label{fig: all_model_concept_acc_middle_zh}
 \vspace{-2mm}
\end{figure*}

\section{Details on the Efficient Fine-Tuning}
\label{app: eft}
 In this section, we provide the details on the efficient fine-tuning to enhance mathematical reasoning abilities towards specific concepts by first training a concept classifier and then curating a set of samples from a large open-sourced math dataset.
Specifically,
first, by additionally collecting {extra 10 problems per concept}, we construct a classifier capable of identifying the concept class of a given question. The backbone of this classifier is a pretrained bilingual LLM (i.e., Baichuan2-13B), where the classification head is operated on its last hidden output feature.
Note that the concept classification accuracies in English and Chinese are 92.5 and 86.9,
respectively,
which indicates that it is reasonable to use an additional classifier for curating an extra concept-related dataset from large-scale math-related data.
Note that in our work, we crawl from the OpenWebMath~\cite{paster2023openwebmath} to produce the concept-related training dataset. 

\section{Details on the Evaluated Models}
In this section, we offer a detailed overview of the Large Language Models (LLMs) and present the corresponding model links in Table~\ref{tab:model_links}.
\label{app: em}
\begin{itemize}[leftmargin=4mm]
    \item GPT-3.5/GPT-4~\cite{gpt4}: The most powerful closed-model from OpenAI. We utilize its API: gpt-3.5-turbo and gpt-4.
    \item LLaMa2-7B/13B/70B~\cite{touvron2023llama2}: A set of open-source models developed by Meta.
    \item Qwen-14B/72B~\cite{qwen}: This model pre-trained on multilingual data, concentrates on Chinese and English languages. We employ both the  Qwen-Base-14B, and the Qwen-Base-72B.
    \item Baichuan2-13B~\cite{baichuan}: This model demonstrates impressive performance in both Chinese and English benchmarks.
    \item MetaMath-13B~\cite{megill2019metamath}: A domain-specific language model for mathematical reasoning, fine-tuned from the LLaMA-2 model using the MetaMathQA~\footnote{\url{https://huggingface.co/datasets/meta-math/MetaMathQA}} dataset.
    \item WizardMath-13B~\cite{luo2023wizardmath}: Another domain-specific language model for mathematical reasoning, fine-tuned from the LLaMA-2 model using reinforcement learning.
    \item MAmmoTH-13B~\cite{yue2023mammoth}: This model is specifically designed for general math problem-solving and has been fine-tuned from the LLaMA model using the MathInstruct~\footnote{\url{https://huggingface.co/datasets/TIGER-Lab/MathInstruct}} dataset. This dataset features training data that includes both chain-of-thought (CoT) and program-of-thought (PoT) rationales.
    \item Yi-6B/34B~\cite{2023yi}: This model released by 01 shows promising performance results in both Chinese and English.
    \item ChatGLM3-6B~\cite{zeng2022glm}: a lightweight and high-performance pre-trained dialogue model released by Zhipu AI in both Chinese and English. 
    \item InternLM-7B/20B~\cite{2023internlm}: A Multilingual Language Model with Progressively Enhanced Capabilities released by InternLM team.
    \item InternLM-Math-7B/20B~\cite{ying2024internlmmath}: Well-performed math reasoning language models.
    \item DeepSeekMath-7B~\cite{deepseek-math}: One powerful mathematical language model released by DeepSeek.
\end{itemize}

\begin{table*}[h]
\resizebox{1.0\textwidth}{!}{
\begin{tabular}{lll}
\toprule
\multicolumn{2}{c}{Models}  & HuggingFace Link / OpenAI Model \\ \midrule
ChatGLM3 & ChatGLM3-6B          & \url{https://huggingface.co/THUDM/chatglm3-6b}                       \\ 
DeepSeekMath & DeepSeekMath-7B  & \url{https://huggingface.co/deepseek-ai/deepseek-math-7b-instruct}   \\ 
Baichuan2 & Baichuan2-13B       & \url{https://huggingface.co/baichuan-inc/Baichuan2-13B-Chat}         \\ 
MetaMath & MetaMath-13B         & \url{https://huggingface.co/meta-math/MetaMath-13B-V1.0}             \\ 
WizardMath & WizardMath-13B     & \url{https://huggingface.co/WizardLM/WizardMath-13B-V1.0}            \\ 
MAmmoTH & MAmmoTH-13B           & \url{https://huggingface.co/TIGER-Lab/MAmmoTH-13B}          \\ \midrule
\multirow{4}{*}{InternLM} & InternLM-7B      & \url{https://huggingface.co/internlm/internlm2-chat-7b}  \\
                         & InternLM-20B      & \url{https://huggingface.co/internlm/internlm2-chat-20b} \\ 
                         & InternLM-Math-7B  & \url{https://huggingface.co/internlm/internlm2-math-7b}  \\
                         & InternLM-Math-20B & \url{https://huggingface.co/internlm/internlm2-math-20b} \\ \midrule
\multirow{2}{*}{Yi}      & Yi-6B             & \url{https://huggingface.co/01-ai/Yi-6B-Chat}             \\
                         & Yi-34B            & \url{https://huggingface.co/01-ai/Yi-34B-Chat}            \\ \midrule
\multirow{3}{*}{LLaMA2}  & LLaMA2-7B         & \url{https://huggingface.co/meta-llama/Llama-2-7b-chat-hf}\\
                         & LLaMA2-13B        & \url{https://huggingface.co/meta-llama/Llama-2-13b-chat-hf}\\
                         & LLaMA2-70B        & \url{https://huggingface.co/meta-llama/Llama-2-70b-chat}   \\ \midrule
\multirow{2}{*}{Qwen}   & Qwen-14B           & \url{https://huggingface.co/Qwen/Qwen-14B-Chat}           \\
                        & Qwen-72B           & \url{https://huggingface.co/Qwen/Qwen-72B-Chat}      \\ \midrule
\multirow{2}{*}{GPT}    & GPT-3.5            & gpt-3.5-turbo \\
                        & GPT-4              & gpt-4      \\ 
\bottomrule
\end{tabular}
}
\caption{Model links.}
\label{tab:model_links}
\vspace{-0.5cm}
\end{table*}

\section{More Results}
\label{app: results}
As shown in Fig.~\ref{fig: mean_concept_acc_element_en} and Fig.~\ref{fig: mean_concept_acc_element_zh},
we provide the mean concept accuracies of Elementary-EN and Elementary-ZH of the evaluated models across different concepts.

As shown in Fig.~\ref{fig: concept_acc_element_en} and Fig.~\ref{fig: concept_acc_element_zh},
we illustrate the concept accuracies on Elementary-EN and Elementary-ZH for different selected models.
For the results of all models,
please refer to Fig.~\ref{fig: all_model_concept_acc_element_en}, 
Fig.~\ref{fig: all_model_concept_acc_middle_en},
Fig.~\ref{fig: all_model_concept_acc_element_zh} and Fig.~\ref{fig: all_model_concept_acc_middle_zh}.

\section{Details on 5-shot Prompts}
\label{app: prompts}
We provide the 5-shot prompts for our ConceptMath in Pages 17-20.

\begin{figure*}[t]{
\label{ele-en-prompts}
\begin{tcolorbox}[colback=white!95!gray,colframe=gray!50!black,rounded corners,label={ele-en-prompts}, title={Elementary-EN (5-shot prompts)}]
\textcolor{blue}{Q}: 
    For a party, Oscar gets 4 different kinds of soda. He buys a 6-pack of each kind. 
    At home, he divides the sodas evenly among 3 coolers. How many sodas are in each cooler?\\
    \textcolor{blue}{A}: Oscar buys 4 different kinds of soda, and he gets a 6-pack of each kind. This means that he has 4 * 6 = 24. sodas in total. He divides these 24 sodas evenly among 3 coolers. So number of sodas per cooler is 24 / 3 = 8. Therefore, the answer is 8.
    \\\textcolor{blue}{Q}:  Amazing Animals Zoo is building a new home for its 8 zebras. Each zebra will get 2 acres of grassy land to roam around in. The zebras will have a pond that covers 1 acre. How big will the zebras' new home be? 
    \\\textcolor{blue}{A}: The total required grassland area for the zebras is: 8 zebras * 2 acres per zebra = 16 acres. Then, we need to add the extra 1 acre for the pond: 16 acres (for the zebras) + 1 acre (for the pond) = 17 acres. Therefore, the answer is  17.
    \\\textcolor{blue}{Q}:  The state taxes personal income exceeding 3500 yuan (this portion not to exceed 3500 yuan) at a rate of 3\%. If Xiaoming's father's monthly salary is 4700 yuan, how much personal income tax should Xiaoming's father pay each month? 
    \\\textcolor{blue}{A}: To calculate the amount of personal income tax Xiaoming's father should pay, we can follow\\this simple process:Subtract the tax-free threshold from the monthly salary.\\Taxable income = Monthly salary - Tax-free threshold = 4700 - 3500 = 1200 yuan. Apply the tax rate to the amount that exceeds the threshold. Tax owed = Taxable income $*$ Tax rate = 1200 * 3\% = 36 yuan.
    Therefore, the answer is 36.
    \\\textcolor{blue}{Q}: Find the 10th term in the arithmetic sequence where the first term is 3 and the common difference is 4. 
    \\\textcolor{blue}{A}: 
    To find the nth term in an arithmetic sequence, we can use the formula: $a_n = a_1 + (n - 1) * d$, 
    where: $a_n$ is the $n$th term, $a_1$ is the first term, d is the common difference, n is the term number.
     Given that the first term ($a_1$) is 3 and the common difference (d) is 4, we can find the $10$th term ($a_{10}$) by plugging these values into the formula: 
     $a_{10} = a_1 + (10 - 1) * d = 3 + 9 * 4 = 3 + 36 = 39$.
    Therefore, the answer is 39.
    \\\textcolor{blue}{Q}:  A regular polygon has an interior angle of 156 degrees. How many sides does the polygon have? 
    \\\textcolor{blue}{A}: 
    To find the number of sides of a regular polygon with a given interior angle, we can use the formula:
    Interior Angle = [(n - 2) * 180] / n, where n is the number of sides. Given that interior angle is 156: 156 = [(n - 2) * 180] / n. Solve for n, so we get n = 15. 
    Therefore, the answer is 15. 
\end{tcolorbox}}
\end{figure*}

\begin{figure*}[t]
\label{mid-en-prompts}
\begin{tcolorbox}[colback=white!95!gray,colframe=gray!50!black,rounded corners,label={mid-en-prompts}, title={Middle-EN (5-shot prompts)}]
\textcolor{blue}{Q}: $g(x) = x^2$, $f(x) = 3(x-2)^2$, $h(x)=g(x)-f(x)$, $y=h(23)$, give me the value of $y$.
    \\\textcolor{blue}{A}: 
    First, calculate $g(23)$ and $f(23)$: $g(23) = 23^2 = 529$.
    Next, calculate $f(23)$: $f(23) = 3 * (23-2)^2 = 3 * (21)^2 = 3 * 441 = 1323$. 
    Now, we calculate $h(23)$: $h(23) = g(23) - f(23) = 529 - 1323 = -794$.
    Therefore, the answer is -794.\\
    \textcolor{blue}{Q}: What is the slope of the line segment between the two points: (3,1), (5,2)?\\
    \textcolor{blue}{A}: 
    The slope of a line between two points $(x_1, y_1)$ and $(x_2, y_2)$ can be found using the formula:  $slope (m) = (y_2 - y_1) / (x_2 - x_1)$. 
    Given the two points (3, 1) and (5, 2), we can substitute these values into the formula: $slope(m) = (2 - 1) / (5 - 3) = 1/2 = 0.5$.
    Therefore, the answer is 1/2.\\
    \textcolor{blue}{Q}: In a party, there are 5 cupcakes remaining and 10 children. A mother inserts 10 pieces of paper
    into a hat, 5 with a cupcake image and 5 without. The children who draw the cupcake paper will 
    receive a cupcake. If the first child draws a paper with a cupcake image, what is the probability
    that the second child will also draw a cupcake paper?
    \\\textcolor{blue}{A}: 
    Initially, there are 5 cupcake papers and 5 non-cupcake papers in the hat, making a total of 10
    papers. After the first child draws a cupcake paper, there are 4 cupcake papers and 5 non-cupcake
    papers left in the hat, with a total of 9 papers. The probability that the second child will draw
    a cupcake paper is then the number of cupcake papers left divided by the total number of papers left:
    Probability = Number of cupcake papers left / Total number of papers left = 4/9. 
    Therefore, the answer is 4/9.\\
    \textcolor{blue}{Q}: 
    What is the total area of an irregular polygon that consists of two squares and two
    rectangles where the sides of the squares are 12 cm and 16 cm respectively, and the
    dimensions of the rectangles are 13 cm by 20 cm and 10 cm by 7 cm respectively?\\
    \textcolor{blue}{A}: 
    To find the total area of an irregular polygon consisting of two squares and two
    rectangles, you would sum up the individual areas of each shape.
    The area of a square is given by the formula  $A = s^2$, where s is the length of
    a side of the square.
    For the first square with a side of 12 cm: $A_1 = 12^2 = 144 cm^2$.
    For the second square with a side of 16 cm: $A_2 = 16^2 = 256 cm^2$.
    The area of a rectangle is given by the formula ( $A = l * w$ ), where $l$ is the 
    length and $w$ is the width of the rectangle.
    For the first rectangle with dimensions 13 cm by 20 cm: $A_3 = 13 * 20 = 260 cm^2$
    For the second rectangle with dimensions 10 cm by 7 cm: $A_4 = 10 * 7 = 70 cm^2$
    $Tota Area = A_1 + A_2 + A_3 + A_4 = 144 + 256 + 260 + 70 = 730 cm^2$
    Therefore, the answer is 730.\\
    \textcolor{blue}{Q}: 
    Susie is making invitations to her party this coming weekend. It takes her about 5
    minutes to make each invitation. It took her 2 hours to finish all the invitations.
    How many people is Susie inviting?\\
    \textcolor{blue}{A}: 
    First, convert the 2 hours into minutes to match the unit of time used for the
    invitation-making process. 2 hours = 120 minutes.
    Then, divide the total time spent by the time it takes to make one invitation to
    find out how many invitations she made:
    120 minutes total / 5 minutes per invitation = 24 invitations
    Therefore, the answer is 24.\\
\end{tcolorbox}
\end{figure*}


\begin{CJK*}{UTF8}{gbsn}
\begin{figure*}[t]
\label{ele-zh-prompts}
\begin{tcolorbox}[colback=white!95!gray,colframe=gray!50!black,rounded corners,label={ele-zh-prompts}, title={Elementary-ZH (5-shot prompts)}]
 \textcolor{blue}{Q}: 
    小文用量角器测量一个角的度数时，误把外圈刻度看成了内圈刻度，量出的角是$65°$，这
    个角实际是多少度？\\
    \textcolor{blue}{A}: 
    量角器一般有两个圈的刻度，内圈刻度通常是顺时针方向从0度到180度，外圈刻度通常是逆
    时针方向从0度到180度。如果小文误把外圈刻度看成了内圈刻度，那么测量出的角度其实是
    从180°减去错误读数。因此，实际的角度应该是：180° - 65° = 115°。
    因此，答案是115。
    \\\textcolor{blue}{Q}: 
    小明以每小时2千米的速度上山，到了山顶后立即以每小时3千米的速度按原路返回。求小明
    上下山的平均速度是多少千米？
    \\\textcolor{blue}{A}: 
    要计算小明上下山的平均速度，我们需要知道他上山和下山所走的总距离以及总时间。
    首先，总距离 = 上山距离+下山距离=d+d= 2d千米。
    然后，计算总时间：上山时间 = 上山距离/上山速度 = d/2小时，下山时间 = 下山距离/下山速度 = d/3小时，因此，总时间=上山时间+下山时间 = d/2 + d/3小时。
    最后，计算平均速度：平均速度=总距离 / 总时间 = 2d/(d/2+d/3)=2d/(5d/6) = 12/5 = 2.4千米/小时。
    因此，答案是2.4千米/小时。
    \\\textcolor{blue}{Q}: 一个礼盒长20cm、宽15cm、高10cm，用彩带捆扎这样的一个礼盒，至少需要多长的彩带？（接头处12厘米）
    \\\textcolor{blue}{A}: 
    要捆扎这样一个礼盒，通常情况下我们会沿着礼盒的长和宽方向各绕一圈，即绕过礼盒的长和宽两次，然后再留一些长度作为接头。
    首先，计算沿着礼盒的长方向绕一圈的长度：绕长方向长度 = （礼盒的长 + 礼盒的高）* 2 = （20 + 10）* 2 = 30 * 2 = 60 cm
    然后，计算沿着礼盒的宽方向绕一圈的长度：绕宽方向长度 = （礼盒的宽 + 礼盒的高）* 2 = （15 + 10）* 2 = 25 * 2 = 50 cm
    最后，计算总需要彩带长度：总长度 = 绕长方向长度 + 绕宽方向长度 + 接头长度 = 60 + 50 + 12 = 122 cm
    因此，答案是122厘米。
    \\\textcolor{blue}{Q}: 千岛湖大桥的一侧每隔10米安装1盏灯，共需91盏（两端都装），现在改装61盏（两端都装），平均每相邻两盏灯之间的距离是多少米？
    \\\textcolor{blue}{A}: 
    原来每隔10米安装1盏灯，共需91盏，这意味着桥的总长度覆盖了90个10米的间隔，所以桥的总长度 = 90间隔 * 10米/间隔 = 900米。
    现在改装61盏灯（两端都装），也就是说将这900米分成60个间隔（因为61个灯包含了两端的灯，所以间隔数比灯的数量少一个），每个间隔的长度 = 900米 / 60间隔 = 15米/间隔。
    因此，答案是115米。
    \\\textcolor{blue}{Q}: a是b的3/4，则a比b少(  )，b比a多(  )，b与a的比是(  )，如果a、b两数的和是210，则a＝(  )，b＝(  )，括号处分别为多少？
    \\\textcolor{blue}{A}: 
    已知a是b的3/4，则a = 3/4  * b。因此，a比b少 (1-3/4)/1 = 1/4，b比a多（1-3/4)/(3/4) = 1/3，b与a的比为b : (3/4 * b) = 4:3
    若已知a、b两数的和是210，则a+b = 210，将a = (3/4) * b 代入方程得到：3/4 * b + b  = 210，解得b = 120，a = 3/4 * b = 3/4 * 120 = 90
    因此，答案分别是 1/4，1/3，4:3，90，120。
\end{tcolorbox}
\end{figure*}
\end{CJK*}

\begin{CJK*}{UTF8}{gbsn}
\begin{figure*}[t]
\label{mid-zh-prompts}
\begin{tcolorbox}[colback=white!95!gray,colframe=gray!50!black,rounded corners,label={mid-zh-prompts}, title={Middle-ZH (5-shot prompts)}]
  \textcolor{blue}{Q}: 求不等式$-x^2-2x+8>0$的解集?
   \\\textcolor{blue}{A}: 
    $-x^2 - 2x + 8 > 0$, 通过分解因式方式，得到 $-(x - 2)(x + 4) > 0$ 因此，答案是 (-4, 2)。\\
    \textcolor{blue}{Q}{: 已知函数$y=2x+3$与的图象$y=-1/2x+m$交y轴于点A、B，两函数图象的交点为P(n,-2)，求m和n的值？}\\
    \textcolor{blue}{A}{: 
    两函数图像交点为P(n,-2)，意味着这个点同时满足上述两个函数方程：
    将P点的坐标(n,-2)代入函数$y=3x+3$，即$-2=3n+3$，解得$n=-5/2$。
    将P点坐标(-5/2,-2)代入函数$y=-1/2x+m$，即$-2=-1/2*(-5/2)+m$，解得m=-13/4。
    因此，答案是m为-13/4，n为-5/2。}\\
    \textcolor{blue}{Q}: 已知PA是圆O的切线，A为切点，PC与圆O相交于B、C两点，PB=2cm，BC=8cm，则PA的长是多少厘米？\\
    \textcolor{blue}{A}: 
    由于PA是圆O的切线，根据切割线定理，我们知道从圆外一点引圆的两条割线，这一点到每条割线与圆的交点的两条线段长的积相等，即$PA^2 = PC * PB$ ,
    已知PB=2cm、BC=8cm，则$PC = PB + BC = 2cm + 8cm = 10cm$。
    $PA^2 = PC * PB  = 10 * 2 = 20$，解得PA = 2$\sqrt{5}$cm。
    因此，答案是2$\sqrt{5}$厘米。\\
    \textcolor{blue}{Q}: 南宋著名数学家杨辉精研数学，著有《详解九章算法》，对数的运算进行了深入研究与总结，类比其中的思想方法，可以解决很多数与式的计算问题已知a，b为实数，且$a+b=3, ab=1$，计算可得：$a^2+b^2=7$，$a^3+b^3=18$，$a^4+b^4=47$，由此求$a^5+b^5$得多少？\\
    \textcolor{blue}{A}: 
    已知 $a + b = 3$和 $ab = 1$，可得：
    $a^2 + b^2 = (a + b)^2 - 2ab = 3^2 - 2 * 1 = 9 - 2 = 7
    a^3 + b^3 = (a + b)(a^2 - ab + b^2) = 3(7 - 1) = 3 * 6 = 18
    a^4 + b^4 = (a^2 + b^2)^2 - 2a^2b^2 = 7^2 - 2 * 1^2 = 49 - 2 = 47
    a^5 + b^5 = (a + b)(a^4 - a^3b + a^2b^2 - ab^3 + b^4) = 3(a^4 + b^4 - a^2b^2) = 3(47 - 1) = 3 * 46 = 138$。
    因此，答案是138。\\
    \textcolor{blue}{Q}: 某单位要招聘1名英语翻译，甲、乙两人报名参加了听、说、读、写等项素质测试，甲的成绩分别为：90分、80分、85分、78分；乙的成绩分别为：78分、82分、85分、88分．如果把听、说、读、写的成绩按$2:2:3:3$计算素质测试平均成绩，那么甲和乙的平均成绩分别是多少？\\
    \textcolor{blue}{A}: 
    甲的平均成绩为 $(90*2 + 80*2 + 85*3 + 78*3) / (2+2+3+3) = (180 + 160 + 255 + 234) / 10 = 829 / 10 = 82.9$ 分，乙的平均成绩为 $(78*2 + 82*2 + 85*3 + 88*3) / (2+2+3+3) = (156 + 164 + 255 + 264) / 10 = 839 / 10 = 83.9$ 分。
    因此，答案是甲的平均成绩是82.9分，乙的平均成绩是83.9分。
\end{tcolorbox}
\end{figure*}
\end{CJK*}

\begin{table*}[!ht]
    \centering
    \setlength\tabcolsep{7.5pt}
    \small
    \begin{tabular}{c|c|c|c}
     \toprule
        \textbf{LEVEL1} & \textbf{LEVEL2} & \textbf{LEVEL3} & \textbf{\# Samples} \\  
        \midrule
        \multirow{17}{*}{\begin{tabular}[c]{@{}c@{}}{\textbf{Calculation \& Properties}}\end{tabular}} &\multirow{13}{*}{\begin{tabular}[c]{@{}c@{}}{Calculation}\end{tabular}} 
                & Add& 19
                \\ ~ & ~ & Decimals &  20
                \\ ~ & ~ & Division &  19
                \\ ~ & ~ & Equations &  18
                \\ ~ & ~ & Fractions &  16
                \\ ~ & ~ & Mixed Operations &  18
                \\ ~ & ~ & Multiple &  18
                \\ ~ & ~ & Numerical Expressions &  20
                \\ ~ & ~ & Place Value &  16 
                \\ ~ & ~ & Powers &  20
                \\ ~ & ~ & Rational Number &  17
                \\ ~ & ~ & Subtraction &  19
                \\ ~ & ~ & Variable Expressions &  19 \\\cmidrule{2-4}
        ~ & \multirow{4}{*}{\begin{tabular}[c]{@{}c@{}}{Properties}\end{tabular}}  
                & Compare & 20 \\  
                ~ & ~ & Count & 18 \\  
                ~ & ~ & Estimation \& Rounding & 20 \\  
                ~ & ~ & Patterns& 19 \\ \midrule
            
        \multirow{12}{*}{\begin{tabular}[c]{@{}c@{}}{\textbf{Geometry}}\end{tabular}}
            & Angles & 17 \\ \cmidrule{2-4}
            ~ & Coordinate Plane & Coordinate Plane & 18 \\  \cmidrule{2-4}
        
        ~ & \multirow{5}{*}{\begin{tabular}[c]{@{}c@{}}{Three-dimensional Shapes}\end{tabular}} 
                & Cones & 17
                \\ ~ & ~ & Cubes & 20
                \\ ~ & ~ & Cylinders & 17
                \\ ~ & ~ & Spheres & 17
                \\ ~ & ~ & Volume of 3D shapes & 18 \\ \cmidrule{2-4}
        ~ & \multirow{5}{*}{\begin{tabular}[c]{@{}c@{}}{Two-dimensional Shapes}\end{tabular}} 
                    & Circles & 17
                \\ ~ & ~ & Perimeter & 19
                \\ ~ & ~ & Polygons & 18
                \\ ~ & ~ & Quadrilaterals & 17
                \\ ~ & ~ & Triangles & 18 \\ \midrule
                
        \multirow{11}{*}{\begin{tabular}[c]{@{}c@{}}{\textbf{Measurement}}\end{tabular}} & 
        \multirow{2}{*}{\begin{tabular}[c]{@{}c@{}}{Basic Knowledge}\end{tabular}} 
                & Temperature& 19 \\  
                ~ & ~ & Time& 20 \\  \cmidrule{2-4}        
        ~ & \multirow{2}{*}{\begin{tabular}[c]{@{}c@{}}{Money}\end{tabular}} 
                & Coin Names \& Value& 17 \\  
                ~ & ~ & Exchanging Money & 17 \\  \cmidrule{2-4}

        ~ & \multirow{3}{*}{\begin{tabular}[c]{@{}c@{}}{Ratio}\end{tabular}} 
                        & Percent& 17 \\  
                    ~ & ~ & Proportion& 18 \\  
                    ~ & ~ & Ratio & 19 \\  \cmidrule{2-4} 
        ~ & \multirow{3}{*}{\begin{tabular}[c]{@{}c@{}}{Size}\end{tabular}}
                    & Area & 19
                    \\ ~ & ~ & Length & 20
                    \\ ~ & ~ & Volume & 20 \\ \cmidrule{2-4}
         ~ & \multirow{1}{*}{\begin{tabular}[c]{@{}c@{}}{Weight}\end{tabular}}
                & Light \& Heavy & 20 \\ 
        \midrule
        \multirow{3}{*}{\begin{tabular}[c]{@{}c@{}}{\textbf{Statistics}}\end{tabular}} & Classifying \& Sorting & Classifying \& Sorting& 17 \\  
        \cmidrule{2-4}
        ~ & Data & Mode/Mean/Median/Range & 19 \\  
                \cmidrule{2-4}

        ~ & Probability& Probability & 16 \\  
           \bottomrule 
    \end{tabular}
        \caption{Details of the hierarchical concepts in Elementary-EN.}
        \label{tab: ele-en}
\end{table*}

\begin{table*}[!ht]
    \centering
    \setlength\tabcolsep{19pt}
    \scriptsize
    \begin{tabular}{c|c|c|c}
     \toprule
        \textbf{LEVEL1} & \textbf{LEVEL2} & \textbf{LEVEL3} & \textbf{\# Samples} \\  
        \midrule
        \multirow{20}{*}{\begin{tabular}[c]{@{}c@{}}\textbf{Calculation}\end{tabular}}  &\multirow{7}{*}{\begin{tabular}[c]{@{}c@{}}{Basic Calculation}\end{tabular}}  
                & Add \& Subtract & 20
                \\ ~ & ~ & Decimals & 19
                \\ ~ & ~ & Divide & 19
                \\ ~ & ~ & Exponents \& Scientific Notation & 16
                \\ ~ & ~ & Fractions \& Decimals & 18
                \\ ~ & ~ & Multiply & 18
                \\ ~ & ~ & Square Roots \& Cube Roots & 20
        \\ \cmidrule{2-4}
                ~ & Consumer Math& Consumer Math & 18
        \\\cmidrule{2-4}
                ~ & Financial Literacy & Financial Literacy& 19
         \\\cmidrule{2-4}
        ~ & \multirow{2}{*}{\begin{tabular}[c]{@{}c@{}}{Integers}\end{tabular}} 
                & Absolute Value & 18 \\  
                ~ & ~ & Opposite Integers& 20
        \\\cmidrule{2-4}
                 ~ & Measurement& Measurement Metric & 19 \\  
                \cmidrule{2-4}
        ~ & \multirow{3}{*}{\begin{tabular}[c]{@{}c@{}}{Number Theory}\end{tabular}} 
                    & Factors & 20 \\  
                 ~  &   ~  & Prime Factorization & 19\\
                 ~  &   ~  & Prime or Composite & 18 
        \\\cmidrule{2-4}
             ~ & Percents & Percents & 20 \\ \cmidrule{2-4}
             ~ & Rational \& Irrational Numbers & Rational \& Irrational Numbers & 18 \\ \cmidrule{2-4}
             ~ & Ratios \& Rates & Proportional Relationships & 18 \\ \cmidrule{2-4}
        ~ & \multirow{2}{*}{\begin{tabular}[c]{@{}c@{}}{Sequences}\end{tabular}}  
                & Arithmetic Sequences & 19 \\  
                ~ & ~ & Geometric Sequences & 18 \\  \midrule
        \multirow{10}{*}{\begin{tabular}[c]{@{}c@{}}\textbf{Expressions,}\\
        \textbf{equations, }\\ \textbf{and functions}\end{tabular}} & \multirow{2}{*}{\begin{tabular}[c]{@{}c@{}}{Equations}\end{tabular}} & Linear Equations & 20 \\  
        ~ & ~ & Systems of Equations & 18 \\  \cmidrule{2-4}
        
        ~ & \multirow{3}{*}{\begin{tabular}[c]{@{}c@{}}{Expressions}\end{tabular}}  & Equivalent Expressions & 20 \\  
          ~ & ~& Radical ~& 17 \\
         ~ & ~& Variable ~& 18 \\
                        \cmidrule{2-4}
        ~ & \multirow{4}{*}{\begin{tabular}[c]{@{}c@{}}{Function}\end{tabular}} & Domain \& Range of Functions & 18 \\
             ~ & ~ & Interpret Functions & 19 \\
             ~ & ~ & Linear Functions & 20 \\
             ~ & ~ & Nonlinear Functions & 18 \\
        \cmidrule{2-4}
        ~ & Inequalities & Inequalities& 19 \\  
        \midrule
        \multirow{16}{*}{\begin{tabular}[c]{@{}c@{}}\textbf{Geometry}\end{tabular}} & Congruence \& Similarity & Congruence \& Similarity & 19 \\  
                                \cmidrule{2-4}
        ~ & \multirow{3}{*}{\begin{tabular}[c]{@{}c@{}}{Coordinate Plane}\end{tabular}}& Axes& 17 \\  
~ & ~ & Distance Between Two Points & 19 \\
~ & ~ & Quadrants & 16 \\
                                \cmidrule{2-4}
        ~ & Scale Drawings & Scale Drawings & 16 \\  
                                \cmidrule{2-4}
        ~ & Slope& Slope & 20 \\  
                                \cmidrule{2-4}
        ~ & \multirow{2}{*}{\begin{tabular}[c]{@{}c@{}}{Three-dimensional Figures}\end{tabular}} 
        & Polyhedra& 19 \\  
        ~ & ~ & Surface Area \& Volume & 17 \\  
                                \cmidrule{2-4}
        ~ & Transformations & Transformations & 18 \\  
                                \cmidrule{2-4}
        ~ & \multirow{7}{*}{\begin{tabular}[c]{@{}c@{}}{Two-dimensional Figures}\end{tabular}}& Circle& 20 \\
~ & ~ & Lines \& Angles & 18 \\
~ & ~ & Perimeter \& Area & 20 \\
~ & ~ & Polygons & 18 \\
~ & ~ & Square & 18\\
~ & ~ & Trapezoids & 16\\
~ & ~ & Triangle & 18\\
\midrule
        \multirow{11}{*}{\begin{tabular}[c]{@{}c@{}}\textbf{Statistic}\\ \textbf{and Probability }\end{tabular}} & \multirow{3}{*}{\begin{tabular}[c]{@{}c@{}}{Data}\end{tabular}} & Center \& Variability& 18 \\  
        ~ & ~ & Mean, Median, Mode \& Range& 19 \\  
        ~ & ~ & Outlier & 20 \\ 
                \cmidrule{2-4}

        ~ & One-variable Statistics& One-variable Statistics & 19 \\  
                \cmidrule{2-4}
        ~ &  \multirow{6}{*}{\begin{tabular}[c]{@{}c@{}}{Probability}\end{tabular}}  & Counting Principle & 16\\
~ & ~  & Independent \& Dependent Events & 16\\
~ & ~  & Make Predictions & 17\\
~ & ~  & Probability of Compound Events & 16\\
~ & ~  & Probability of One Event & 17\\
~ & ~  & Probability of Simple and Opposite Events & 19 \\ 
        \cmidrule{2-4}
        ~ & Two-variable Statistics & Two-variable Statistics & 18 \\  
        \bottomrule
    \end{tabular}
    \caption{Details of the hierarchical concepts in Middle-EN.}
    \label{tab: mid-en}
\end{table*}

\begin{CJK*}{UTF8}{gbsn}
\begin{table*}[!ht]
    \centering
\setlength\tabcolsep{5pt}
\resizebox{1\linewidth}{!}{
\small
    \begin{tabular}{c|c|c|c}
     \toprule
        \textbf{LEVEL1} & \textbf{LEVEL2} & \textbf{LEVEL3} & \textbf{\# Samples} \\ 
        \midrule
        \multirow{13}{*}{\begin{tabular}[c]{@{}c@{}}\textbf{几何}\\ \textbf{(Geometry)}\end{tabular}} & \multirow{4}{*}{三角形(Triangle)} & 全等三角形(Congruent Triangle) & 20 \\  
        ~ & ~ & 勾股定理(Pythagorean Theorem) & 20 \\  
        ~ & ~ & 等腰三角形(Isosceles Triangle) & 20 \\  
        ~ & ~ & 等边三角形(Equilateral Triangle) & 20 \\  
        \cmidrule{2-4}
        ~ & \multirow{2}{*}{四边形(Quadrilateral)} & 平行四边形(Parallelogram) & 20 \\  
        ~ & ~ & 梯形(Trapezium) & 20 \\ 
                \cmidrule{2-4}

        ~ & \multirow{6}{*}{圆(Circle)} & 圆周角(Angle of Circumference) & 20 \\  
        ~ & ~ & 圆心角(Angle of Center) & 20 \\  
        ~ & ~ & 垂径定理(Vertical Path Theorem) & 20 \\  
        ~ & ~ & 弧长和扇形面积(Arc length \& Sector Area) & 20 \\  
        ~ & ~ & 正多边形和圆(Regular Polygons \& Circles) & 20 \\  
        ~ & ~ & 点线圆位置关系(Relations of Point, Line \& Circle) & 20 \\  
                        \cmidrule{2-4}
        ~ & \begin{tabular}[c]{@{}c@{}}{立体图形}\\ {(Three-dimensional Shapes)}\end{tabular} & 圆锥(Cone) & 20 \\  
        \midrule
        \multirow{12}{*}{\begin{tabular}[c]{@{}c@{}}\textbf{函数}\\ \textbf{(Function)}\end{tabular}} & \multirow{5}{*}{一次函数(Linear Function)} & \begin{tabular}[c]{@{}c@{}}{函数与一元一次方程}\\ {(Univariate Function \& Equation)}\end{tabular}& 20 \\  
        ~ & ~ & \begin{tabular}[c]{@{}c@{}}{函数与一元一次不等式 }\\ {(Linear Functions \& Univariate Linear Inequalities)}\end{tabular}& 20 \\  
        ~ & ~ & \begin{tabular}[c]{@{}c@{}}{一次函数与二元一次方程组}\\ {(Linear Functions \& System of Binary Linear Equations)}\end{tabular} & 20 \\  
        ~ & ~ & 正比例函数(Proportional Function) & 20 \\  
        ~ & ~ & \begin{tabular}[c]{@{}c@{}}{一次函数解析式}\\ {(Analytical Formula of Linear Functions )}\end{tabular}& 20 \\  
        \cmidrule{2-4}
        ~ & \multirow{2}{*}{二次函数(Quadratic Function)} & \begin{tabular}[c]{@{}c@{}}{二次函数的应用}\\ {(Applications of Quadratic Functions)}\end{tabular} & 20 \\
        ~ & ~ & \begin{tabular}[c]{@{}c@{}}{抛物线的性质}\\ {(Properties of Parabolas)}\end{tabular} & 18 \\  
                        \cmidrule{2-4}

        ~ & \multirow{3}{*}{\begin{tabular}[c]{@{}c@{}}{反比例函数}\\ {(Inverse Proportional Function)}\end{tabular}} & 定义(Definition) & 20 \\  
        ~ & ~ & 应用(Applications) & 20 \\  
        ~ & ~ & 性质(Properties) & 19 \\ 
                        \cmidrule{2-4}
        ~ & \multirow{2}{*}{\begin{tabular}[c]{@{}c@{}}{平面直角坐标系}\\ {(Rectangular Coordinate System)}\end{tabular}} & 有序数对(Ordered Pair) & 20 \\  
        ~ & ~ & 象限中的点(Points of Quadrant) & 14 \\  
\midrule
       \multirow{16}{*}{\begin{tabular}[c]{@{}c@{}}\textbf{数与式}\\ \textbf{(Number}\\ \textbf{and Expression)}\end{tabular}} & \multirow{2}{*}{代数式(Algebra Expression)} & 代数式求值(Algebraic Expression Evaluation) & 20 \\  
        ~ & ~ & 同类项(Similar Items) & 20 \\  
                        \cmidrule{2-4}

        ~ & \multirow{2}{*}{分式(Fraction)} & 指数幂(Exponential Power) & 20 \\  
        ~ & ~ & 约分(Fraction Reduction) & 19 \\ 
                        \cmidrule{2-4}

        ~ & \multirow{2}{*}{因式(Factor)} & 十字相乘法(Cross Multiplication) & 20 \\  
        ~ & ~ & 公因式提取(Common Factor Extraction) & 18 \\  
                        \cmidrule{2-4}

        ~ & \multirow{2}{*}{应用(Application)} & 流水问题(Flow Problem) & 20 \\  
        ~ & ~ & 鸽巢问题(Pigeon Nest Problem) & 20 \\  
                        \cmidrule{2-4}

        ~ & \multirow{3}{*}{整式(Integral Expression)} & 乘法公式(Multiplication) & 20 \\  
        ~ & ~ & 整式的乘除及混合(Multiplication, Division \& Mixing) & 20 \\  
        ~ & ~ & 整式的加减(Addition \& Subtraction) & 20 \\ 
                        \cmidrule{2-4}

        ~ & 无理数(Irrational Number) & 无理数识别(Irrational Number Recognition) & 20 \\  
                        \cmidrule{2-4}

        ~ & \multirow{4}{*}{根式(Radical Expression)} & 二次根式的运算(Operation of Quadratic Radicals) & 20 \\  
        ~ & ~ & 同类二次根式(Similar Quadratic Radicals) & 20 \\  
        ~ & ~ & 平方根与算术平方根(Square Root \& Arithmetic Square Root) & 20 \\  
        ~ & ~ & 立方根(Cube Root) & 20 \\  
\midrule
        \multirow{11}{*}{\begin{tabular}[c]{@{}c@{}}\textbf{方程与不等式}\\ \textbf{(Equations \& Inequalities)}\end{tabular}} & 
        \multirow{2}{*}{\begin{tabular}[c]{@{}c@{}}{一元一次方程}\\ {(Linear Equation in One Variable)}\end{tabular}}& 一元一次方程的应用(Applications) & 20 \\  
        ~ & ~ & 解一元一次方程(Solutions) & 20 \\  
                                \cmidrule{2-4}

        ~ & \multirow{2}{*}{\begin{tabular}[c]{@{}c@{}}{一元二次方程}\\ {(Quadratic Equation in One Variable)}\end{tabular}} & 一元二次方程的应用(Applications) & 20 \\  
        ~ & ~ & 解一元二次方程(Solutions) & 20 \\  
                                \cmidrule{2-4}

        ~ & \multirow{4}{*}{\begin{tabular}[c]{@{}c@{}}{不等式与不等式组}\\ {(Inequalities \& Groups of Inequalities)}\end{tabular}} & 
         \multirow{1}{*}{\begin{tabular}[c]{@{}c@{}}{一元一次不等式的应用} {(Applications of Unary First Order Inequality)}\end{tabular}}& 20 \\ 
        ~ & ~ & 一元一次不等式组的应用(Applications of Unary First Order Groups of Inequalities) & 20 \\  
        ~ & ~ & 解一元一次不等式(Solve the First Inequality of One Variable) & 20 \\  
        ~ & ~ & 解一元一次不等式组(Solve Unary First Order Groups of Inequalities) & 20 \\  
                                \cmidrule{2-4}

        ~ & \multirow{2}{*}{分式方程(Fractional Equation)} & 分式方程的应用(Application of Fractional Equation) & 20 \\  
        ~ & ~ & 解分式方程(Solve Fractional Equation) & 20 \\  
\midrule
        \multirow{5}{*}{\begin{tabular}[c]{@{}c@{}}\textbf{统计与概率}\\ \textbf{(Statistics and} \\ \textbf{ Probability)}\end{tabular}} & \multirow{2}{*}{数据分析(Data Analysis)} & 数据的波动趋势(Fluctuating Trend of Data) & 20 \\  
        ~ & ~ & 数据的集中趋势(Central Tendency of Data) & 20 \\  
                                \cmidrule{2-4}
        ~ & \multirow{3}{*}{概率(Probability)} & 概率的应用(Applications of Probability) & 20 \\  
        ~ & ~ & 求概率(Find Probability) & 20 \\  
        ~ & ~ & 随机事件与概率(Random Events \& Probabilities) & 20 \\  
        \bottomrule
    \end{tabular}}
            \caption{Details of the hierarchical concepts in Middle-ZH.}
            \label{tab: mid-zh}
\end{table*}
\end{CJK*}

\end{document}